\documentclass[11pt,a4paper]{article}
\usepackage[hyperref]{eacl2021}
\usepackage{times}
\usepackage{latexsym}

\usepackage{lipsum}

\usepackage{microtype}

\aclfinalcopy 

\usepackage{amsmath,amsfonts,bm}

\def\eqref#1{equation~\ref{#1}}

\def\1{\bm{1}}

\def\va{{\bm{a}}}

\def\ve{{\bm{e}}}

\def\vk{{\bm{k}}}

\def\vo{{\bm{o}}}

\def\vq{{\bm{q}}}

\def\vv{{\bm{v}}}

\def\vx{{\bm{x}}}
\def\vy{{\bm{y}}}

\DeclareMathAlphabet{\mathsfit}{\encodingdefault}{\sfdefault}{m}{sl}
\SetMathAlphabet{\mathsfit}{bold}{\encodingdefault}{\sfdefault}{bx}{n}

\newcommand{\R}{\mathbb{R}}

\usepackage[pdftex]{graphicx}
\usepackage{caption}
\usepackage{subcaption}
\usepackage{float}

\usepackage{pgfplots}
\pgfplotsset{compat=newest}
\usepgfplotslibrary{groupplots}
\usepgfplotslibrary{dateplot}

\usepackage{pdflscape}

\def \mainPlotHeight {5cm}

\makeatletter
\newcommand*{\centerfloat}{%
  \parindent \z@
  \leftskip \z@ \@plus 1fil \@minus \marginparwidth
  \rightskip \leftskip
  \parfillskip \z@skip}
\makeatother

\title{Telling BERT's Full Story: from Local Attention to Global Aggregation}

\author{Dami\'an Pascual, Gino Brunner and Roger Wattenhofer \\
Department of Electrical Engineering and Information Technology \\
ETH Zurich, Switzerland \\
\{dpascual,brunnegi,wattenhofer\}@ethz.ch
}

\date{}

\begin{document}
\maketitle
\begin{abstract}
We take a deep look into the behaviour of self-attention heads in the transformer architecture. In light of recent work discouraging the use of attention distributions for explaining a model's behaviour, we show that attention distributions can nevertheless provide insights into the local behaviour of attention heads. This way, we propose a distinction between local patterns revealed by attention and global patterns that refer back to the input, and analyze BERT from both angles. We use gradient attribution to analyze how the output of an attention head depends on the input tokens, effectively extending the local attention-based analysis to account for the mixing of information throughout the transformer layers. We find that there is a significant mismatch between attention and attribution distributions, caused by the mixing of context inside the model. We quantify this discrepancy and observe that interestingly, there are some patterns that persist across all layers despite the mixing.
\end{abstract}

\section{Introduction}\label{intro}

The inception of the transformer architecture has sparked significant progress across a wide range of language understanding tasks. Variants of transformers currently dominate the popular GLUE~\cite{GLUE2019} and SuperGLUE~\cite{SuperGLUE2019} benchmarks and have even achieved super human performance on multiple tasks.
The main innovations behind the transformer architecture are the stacking of self-attention layers into a multi-layer self-attention architecture, as well as an unsupervised pre-training phase that primes the model to be fine-tuned on a wide range of language tasks. 
Transformers and other self-attention-based models have been successfully adopted in other areas such as computer vision~\cite{image}, music processing~\cite{music} or protein research~\cite{protein}. 
Their extraordinary empirical success has led researchers to investigate transformers in order to better understand the source of this success, but also in an attempt to explain model decisions.

Much of the research around interpretability and explainability is focused on analyzing the self-attention operation~\cite{whatBertLooksAt}.
In multi-layer self-attention, every input computes an attention distribution over itself and all other inputs to produce ever more complex feature representations. In the case of language, a word in a sentence attends to itself and to all other words in order to compute an updated contextual representation of itself. 
It is tempting to directly rely on attention distributions to explain the model's predictions. The rationale is that if the attention distribution aligns with human intuition, we can conclude that the model learned robust features and obtained a deep understanding of language, in contrast to simply overfitting on spurious patterns. For example, if a transformer classifies an online comment as hate speech, but we find that the model mostly attended to neutral or even positive words, we would conclude that the model did not actually understand the text and that the correct prediction was either due to chance or to the exploitation of an underlying statistical bias in the data \cite{bertstatcue2019}.

However, recent studies \cite{Brunner2020On,DeceiveAttention2019} question the ability of attention maps to provide a faithful explanation of the inner workings of transformer models. In particular, when the explanations refer to the model input, attention maps do not account for the mixing of information throughout the model. Since self-attention mixes information among all input tokens, the hidden layers attend over mixtures of tokens. Therefore, attention maps may be useful to investigate the local behavior of attention heads but not to draw conclusions about how input tokens relate to each other. 

In this work we take a detailed look at the inner workings of BERT's attention heads, both by analyzing the self-attention distributions, and by using gradient attribution to account for the mixing of tokens throughout the model.
We first show that self-attention distributions correlate strongly with Hidden Token Attribution~\cite{Brunner2020On} (HTA) from hidden embedding to head output, this result validates HTA. We then present novel location based attention patterns, revealing that BERT, despite its bi-directional language modeling objective, attends to past embeddings in earlier layers, and to future ones in later layers. Next, we use HTA in order to extend the analysis to take the mixing of information into account, which allows to draw conclusions about the behaviour of an attention head with respect to the original input word. The patterns that emerge are different from the local attention-based patterns, giving us deeper insight into the operation of the model and emphasizing that local attention-based explanations are very different from global attribution-based explanations. 
Finally, we contrast attention and HTA distributions for individual examples. Our results further highlight the discrepancy between local attention patterns and global attribution patterns.

\section{Related Work}

The good performance of attention \cite{graves2013generating, originalattentionpaper} models in Natural Language Processing (NLP) arises from their ability to learn alignments between words. The transformer architecture \cite{Attentionisallyouneed2017} is a multi-layer multi-head self-attention architecture that is pre-trained in an unsupervised manner. The extraordinary performance of transformer models has accelerated progress in the field of NLP. Currently, there is a growing number of different transformer models that vary in size, pre-training objective and/or other architectural elements \cite{GPT, GPT2, roberta, albert, xlnet, distilbert, reformer, t5}. 

The success of transformers and the possibility of visualizing attention distributions \cite{Attentionisallyouneed2017}, has motivated a line of research aiming to understand the inner workings of transformers and explain their decisions. Many of these studies have focused on BERT~\cite{BERT}, a well-known transformer model, leading to a body of research grouped under the term BERTology~\cite{bertology}. 

Aforementioned research builds on previous work on the interpretability of attention distributions in other models apart from transformers. In particular, \citet{attentionIsNotExplanation} examine the attention distributions of LSTM based encoder-decoder models and show a weak to moderate correlation between attention and dot-product gradient attribution. Furthermore, they show that adversarial attention distributions that do not change the model's decision can be constructed.
In the same line, \citet{isattentioninterprable} find, through zeroing out attention weights, that gradient attribution is a better predictor of feature importance with respect to the model's output than attention weights.
\citet{attentionisNotNotexplanation} find that although adversarial attention distributions can be easily obtained, they perform worse on a simple diagnostic task.
All of these works raise concerns about the ability of attention distributions to explain the decisions of a model.

Despite existing concerns surrounding the interpretability of attention distributions, very few works have studied how this problem affects transformers. 
\citet{DeceiveAttention2019} show that, just as in other attention models, it is possible to manipulate self-attention in transformers in order to generate different attention masks that cause only a small drop in performance. 
\citet{Brunner2020On} find that attention distributions are not unique when the sequence length is larger than the head dimension and show that this can lead to the discovery of spurious patterns. Furthermore, they show that although it is possible to map hidden tokens back to their corresponding input tokens, there is a very large degree of information mixing inside the model, which raises questions about straightforward interpretations of attention maps. Recently, \cite{abnar2020zuidema} proposed a method to quantify information flow inside transformers. This method tracks the mixing of information due to attention but omits the effect of feed-forward networks.

Our work addresses this important issue by distinguishing between local and global aggregation patterns, where the former can be explained by attention distributions and the latter by attribution. We analyze BERT from both angles and quantify the mismatch between these interpretations. We show that attention correlates well with attribution locally but not globally and therefore attention maps are inadequate to draw conclusions that refer to the input of the model.

\section{Background on Transformers}

The original transformer architecture~\cite{Attentionisallyouneed2017} is a sequence-to-sequence model consisting of an encoder and a decoder, both of which follow a multi-layer multi-head self-attention structure. Conversely, most of the pre-trained transformer models that can be fine-tuned on supervised language understanding tasks only consist of an encoder.
Each transformer layer consists of a self-attention block and a non-linear feed forward block (MLP) with layer normalizations~\cite{ba2016layer}. 

The input to a transformer layer is a sequence of embeddings $E^l = [\ve_0^l, ..., \ve_{d_s}^l] \in \R^{d_e \times d_s}$, where $l$ denotes the layer index, $d_e$ is the embedding dimension, and $d_s$ is the sequence length. We refer to the sequence of non-contextual input word embeddings as $E^0$, and to the hidden contextual embeddings as $E^l$, where $l>0$. Note that $E^0$ refers to the word embeddings after position and sequence embeddings have been added.
A self-attention block consists of $n_h$ separate attention heads. The attention heads independently perform the self-attention operation, and the results are then concatenated and projected back into the embedding space by a linear layer. The output of the attention block is then fed into the MLP.

The self-attention operation itself is implemented by projecting each input token $\ve_i \in \R^{d_e}$ into a query vector $\vq_i \in \mathbb{R}^{d_q}$, key vector $\vk_i \in \mathbb{R}^{d_q}$ and value vector $\vv_i \in \mathbb{R}^{d_v}$. We present the self-attention operation from the perspective of a single token $\ve_i$ attending to all input tokens. For that, the key vectors $\vk_i$ are aggregated into the key matrix $K=[\vk_0,...,\vk_{d_s}]\in \R^{d_q \times d_s}$ and the value vectors $\vv_i$ are aggregated into the value matrix $V=[\vv_0,...,\vv_{d_s}] \in \R^{d_v \times d_s}$. The attention distribution $\va_i$ of token $\ve_i$ over all input tokens is then computed as 

\begin{equation*}
    \va_i = \text{softmax}\left(\frac{\vq_i^T \cdot K}{\sqrt{d_q}}\right)
\end{equation*}

The attention vector $\va_i\in \R^{d_s}$ now contains an attention weight for each input token. 
$\va_i$ is then multiplied with the value matrix $V$ to compute the output of the self-attention operation for a token $i$ and a head $h$ as

\begin{equation*}
    \vo_{h,i} =  V \cdot \va_i 
\end{equation*}

The outputs of all heads $\{\vo_{0,i},...,\vo_{n_h,i}\}\in \R^{d_e}$ are then concatenated and fed through a linear layer to compute the output of the self-attention block for a single token. This linear layer can be thought of as an aggregation operation that projects the output of the independent heads back into embedding space. 
In practice, the attention distributions for all tokens are computed in parallel.

\section{Extending Hidden Token Attribution}\label{method}

Hidden Token Attribution \cite{Brunner2020On} is a gradient-based attribution method that quantifies how much information from each input token is contained in a given hidden embedding. For each layer $l$, this method defines the relative contribution $c^l_{i,j}$ of an input token $\ve_i^0$ to a hidden embedding $\ve_j^l$ as: 

\begin{equation}\label{eqn:hta}
c_{i,j}^l = \frac{||\nabla_{i,j}^l||_2}{\sum_{k=0}^{d_s}||\nabla_{k,j}^l||_2}\quad \text{ with}\quad \nabla_{i,j}^l =\frac{\partial \ve_j^l}{\partial \ve_i^0}
\end{equation}

The contribution $c^l_{i,j}$ is normalized by the sum of the attribution values to all input tokens and hence, ranges between 0 and 1. 

In this work, we apply Hidden Token Attribution to the individual attention heads of BERT. For a token $\ve_j^l$ at layer $l$ we back-propagate the gradients from the output $\vo_{h,j}^l$ of each attention head $h$ independently. This differs from the original method in that Hidden Token Attribution propagates the gradients from the layer output.
In general, using Equation~\ref{eqn:hta}, we can compute the contribution between any two vectors in the model, as long as they are connected in the computation graph. We hence denote the contribution of any vector $\vx$ to another vector $\vy$ as $C(\vx,\vy)$.

In particular, we calculate two different contributions to the head output:
\begin{description}
    \item [Previous layer contribution:] Contribution from the hidden embeddings at the input of the attention head to the output of the attention head: $C(\ve_i^{l-1},\vo_{h,j}^l)$
    \item [Input contribution:] Contribution from tokens at the input of the transformer model to the output of an attention head $h$ at layer $l$: $C(\ve_i^0, \vo_{h,j}^l)$ 
\end{description}

\emph{Previous layer contribution} allows us to study how attention heads operate locally and how HTA distributions compare to attention distributions. \emph{Input contribution} enables us to extend the head attention patterns all the way back to the input, thereby controlling for the effect of information mixing.

\section{Setup}\label{setup}

For our experiments we use the non-finetuned, uncased BERT base model~\cite{bert2019} as provided in the original repository.\footnote{https://github.com/google-research/bert}
Despite the recent explosion of new transformer variants, BERT remains the most popular model for research into the interpretability of transformer models. 
The reason for this is that most of the newer models are architecturally similar to BERT, and therefore, studies carried out on BERT either are likely to generalize to these models or can be repeated with relatively little effort.

We perform our experiments on 1800 examples from the development set of the MNLI matched (MNLIm) dataset. 
\citet{Brunner2020On} show that when the sequence length $d_s$ is larger than the head output dimension $d_v$, the attention distributions are not identifiable. Therefore, to guarantee that in our experiments we do not find spurious patterns that do not influence downstream parts of the model, we restrict the examples in our dataset to sequences of maximum length of 64 tokens, which is the head dimension of BERT. Thus, the examples in our dataset have sequence lengths ranging between 6 and 64 tokens, with a median length of 34 tokens. In total, this subset contains 63,456 tokens.

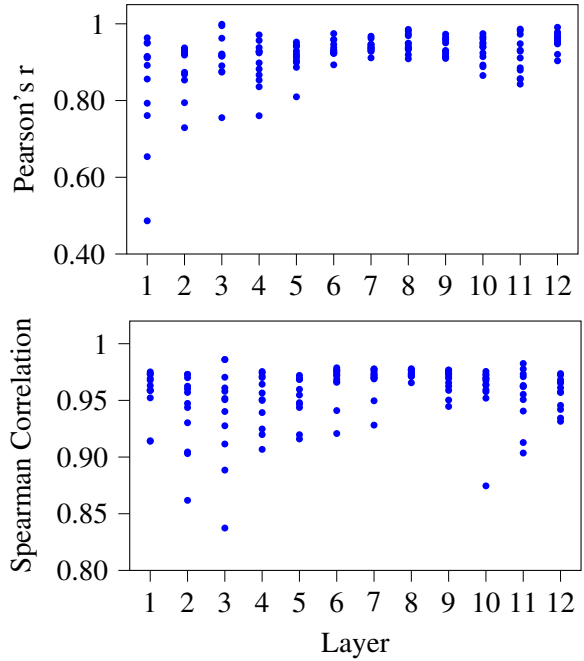
\begin{figure}[t]
\centering
\begin{subfigure}[t]{.47\textwidth}
\centerfloat
\begin{tikzpicture}[scale=1]

\begin{axis}[
tick align=outside,
tick pos=left,
x grid style={white!69.01960784313725!black},
xmin=-0.55, xmax=11.55,
xtick style={color=black},
xtick={0,1,2,3,4,5,6,7,8,9,10,11},
xticklabels={1,2,3,4,5,6,7,8,9,10,11,12},
ytick={0.4, 0.6, 0.8, 1},
yticklabels={0.40,0.60,0.80,1},
y grid style={white!69.01960784313725!black},
ylabel={Pearson's r},
ymin=0.4, ymax=1.05,
ytick style={color=black},
height=0.65\textwidth,
width=\textwidth
]
\addplot [semithick, blue, mark=*, mark size=1, mark options={solid}, only marks]
table {%
0 0.914916977980203
1 0.868258297576141
2 0.995124364660965
3 0.971128080807635
4 0.899214500239615
5 0.925661045303313
6 0.93899800147984
7 0.919486456291142
8 0.965377968716412
9 0.951404179525199
10 0.855890079916741
11 0.965855755705807
};
\addplot [semithick, blue, mark=*, mark size=1, mark options={solid}, only marks]
table {%
0 0.891758186872757
1 0.937243388614364
2 0.915978165459638
3 0.853359107251196
4 0.916008889593689
5 0.974566111850872
6 0.932898474100009
7 0.948677573884893
8 0.930464245887298
9 0.938977897945753
10 0.982235124183028
11 0.950386099705579
};
\addplot [semithick, blue, mark=*, mark size=1, mark options={solid}, only marks]
table {%
0 0.963538632454109
1 0.853435916047973
2 0.890639380901975
3 0.956542188326056
4 0.940584730344604
5 0.923085434278762
6 0.937222132363719
7 0.949568745533454
8 0.950258678762109
9 0.914368294516385
10 0.928648817344652
11 0.920408368738536
};
\addplot [semithick, blue, mark=*, mark size=1, mark options={solid}, only marks]
table {%
0 0.949524687232657
1 0.927114130233791
2 0.755351288195665
3 0.898141774342267
4 0.952383649352812
5 0.924774000110625
6 0.962221021447411
7 0.973861961804918
8 0.924797700162005
9 0.947472650418892
10 0.885354161068898
11 0.966240184855274
};
\addplot [semithick, blue, mark=*, mark size=1, mark options={solid}, only marks]
table {%
0 0.48648853373447
1 0.923364800389607
2 0.916708638134371
3 0.835920027949958
4 0.809718815084502
5 0.893044055483321
6 0.942491812645626
7 0.937096768188473
8 0.958029908493826
9 0.96081146483913
10 0.931212161389372
11 0.947628912977837
};
\addplot [semithick, blue, mark=*, mark size=1, mark options={solid}, only marks]
table {%
0 0.856020694085169
1 0.873648297550903
2 0.875586168553252
3 0.928208526250632
4 0.903805727623169
5 0.958726582093751
6 0.928239499640862
7 0.908843196641061
8 0.959940215817486
9 0.888228522982172
10 0.986200309956374
11 0.957771744703586
};
\addplot [semithick, blue, mark=*, mark size=1, mark options={solid}, only marks]
table {%
0 0.653778399542223
1 0.925188270151801
2 0.962478235673697
3 0.938036469406124
4 0.886576271490183
5 0.959027069826152
6 0.929390066241005
7 0.981601110017896
8 0.972976039426866
9 0.966239885757499
10 0.948102155767997
11 0.95883842308193
};
\addplot [semithick, blue, mark=*, mark size=1, mark options={solid}, only marks]
table {%
0 0.760866062972685
1 0.932536935994957
2 0.874092792239869
3 0.760415065660532
4 0.945563549296667
5 0.923150187788018
6 0.911279787029978
7 0.985459513379629
8 0.919398924394038
9 0.940935164385948
10 0.857121233012901
11 0.99091471847012
};
\addplot [semithick, blue, mark=*, mark size=1, mark options={solid}, only marks]
table {%
0 0.911501727907344
1 0.794423305293785
2 0.92133000211477
3 0.882099676614209
4 0.942751019420736
5 0.937749703321805
6 0.967641897403262
7 0.941058301432426
8 0.90982785841183
9 0.892565312224324
10 0.842777721354186
11 0.978015059496701
};
\addplot [semithick, blue, mark=*, mark size=1, mark options={solid}, only marks]
table {%
0 0.793030039556603
1 0.729277665619554
2 0.998795129738802
3 0.924210619872301
4 0.911678792969226
5 0.929760631503504
6 0.934891441318306
7 0.936008009711573
8 0.928870508924592
9 0.974391837309535
10 0.972870402359351
11 0.903531725166062
};
\addplot [semithick, blue, mark=*, mark size=1, mark options={solid}, only marks]
table {%
0 0.950868462112159
1 0.918517821994164
2 0.920307388738682
3 0.866690983897698
4 0.921343936272006
5 0.931933165616881
6 0.946191237302604
7 0.932736746179925
8 0.914471098970783
9 0.923864386398095
10 0.911041474629746
11 0.958601047564184
};
\addplot [semithick, blue, mark=*, mark size=1, mark options={solid}, only marks]
table {%
0 0.914031018499155
1 0.92305558546132
2 0.916277395205653
3 0.927769554330738
4 0.929159369829877
5 0.946272999270522
6 0.946907347339533
7 0.968877932086224
8 0.912960549060167
9 0.865229420927902
10 0.879290592438682
11 0.974035142561755
};
\end{axis}

\end{tikzpicture}
\end{subfigure}
\hfill
\begin{subfigure}[t]{.47\textwidth}
\centerfloat
\begin{tikzpicture}[scale=1]

\begin{axis}[
tick align=outside,
tick pos=left,
x grid style={white!69.01960784313725!black},
xlabel={Layer},
xmin=-0.55, xmax=11.55,
xtick style={color=black},
xtick={0,1,2,3,4,5,6,7,8,9,10,11},
xticklabels={1,2,3,4,5,6,7,8,9,10,11,12},
y grid style={white!69.01960784313725!black},
ylabel={Spearman Correlation},
ymin=0.8, ymax=1.02,
ytick style={color=black},
ytick={0.8, 0.85, 0.9, 0.95, 1},
yticklabels={0.80, 0.85, 0.90, 0.95, 1},
height=0.65\textwidth,
width=\textwidth
]
\addplot [semithick, blue, mark=*, mark size=1, mark options={solid}, only marks]
table {%
0 0.958824694760993
1 0.930260511515865
2 0.986059010599007
3 0.975460261757571
4 0.946686586858349
5 0.976897173258262
6 0.969818477577378
7 0.971488096433619
8 0.975045935735952
9 0.975550408155858
10 0.940577772122517
11 0.942005921414622
};
\addplot [semithick, blue, mark=*, mark size=1, mark options={solid}, only marks]
table {%
0 0.967851695130303
1 0.972946725885111
2 0.960988767066539
3 0.924766251750907
4 0.959765824260554
5 0.975855944849715
6 0.949644883378625
7 0.973779425860153
8 0.944619052042128
9 0.963975863757157
10 0.977682287134997
11 0.934400455572349
};
\addplot [semithick, blue, mark=*, mark size=1, mark options={solid}, only marks]
table {%
0 0.975125882264224
1 0.903045628543655
2 0.888494596999945
3 0.950656931672532
4 0.91970697599857
5 0.920771117046862
6 0.974724741766138
7 0.976566016826698
8 0.958839792963112
9 0.959388306292545
10 0.963127765428166
11 0.966596980563167
};
\addplot [semithick, blue, mark=*, mark size=1, mark options={solid}, only marks]
table {%
0 0.97294003858263
1 0.94732096076768
2 0.837382773755408
3 0.964266981784143
4 0.972056623303921
5 0.965929578607256
6 0.977458509537592
7 0.977932205305069
8 0.976588184148323
9 0.973190724979745
10 0.90355111392963
11 0.965259649988639
};
\addplot [semithick, blue, mark=*, mark size=1, mark options={solid}, only marks]
table {%
0 0.914358252571249
1 0.972960577401119
2 0.95060680888739
3 0.906776235514391
4 0.91595310403903
5 0.941008723108722
6 0.972993970858318
7 0.971112981506457
8 0.973698036570475
9 0.971897425997156
10 0.962078750462781
11 0.961121321762853
};
\addplot [semithick, blue, mark=*, mark size=1, mark options={solid}, only marks]
table {%
0 0.952237909123824
1 0.943702869693517
2 0.95170588186814
3 0.970180139887144
4 0.968610292760654
5 0.97405038313931
6 0.971059547413088
7 0.965522671978227
8 0.976835964546201
9 0.951959210812818
10 0.962557134185365
11 0.968202884427926
};
\addplot [semithick, blue, mark=*, mark size=1, mark options={solid}, only marks]
table {%
0 0.958702205817222
1 0.970016548703872
2 0.970418112467998
3 0.956517556231061
4 0.947988753098495
5 0.972521870479961
6 0.972468759966516
7 0.976656322574144
8 0.970923419330876
9 0.969559778139161
10 0.970904780200936
11 0.931512923059511
};
\addplot [semithick, blue, mark=*, mark size=1, mark options={solid}, only marks]
table {%
0 0.958856674172798
1 0.957022731581542
2 0.9276040059625
3 0.919850920019782
4 0.968744758264201
5 0.968662094510567
6 0.928175150147628
7 0.972380289016417
8 0.962460147365826
9 0.96909876999273
10 0.912812403415127
11 0.957047829421959
};
\addplot [semithick, blue, mark=*, mark size=1, mark options={solid}, only marks]
table {%
0 0.969307021672497
1 0.904502970433249
2 0.940209055135936
3 0.939246539675138
4 0.954738239581258
5 0.966635739475886
6 0.970005083438969
7 0.976688194645983
8 0.950314429596094
9 0.967395659083703
10 0.955312548561393
11 0.97363154659043
};
\addplot [semithick, blue, mark=*, mark size=1, mark options={solid}, only marks]
table {%
0 0.913946213260414
1 0.861804873836873
2 0.98598164894477
3 0.971170645400142
4 0.943733179174201
5 0.978750287065046
6 0.977837009383133
7 0.971634679590343
8 0.972920685321988
9 0.957665532012766
10 0.973203935212668
11 0.972347040285323
};
\addplot [semithick, blue, mark=*, mark size=1, mark options={solid}, only marks]
table {%
0 0.973392980232048
1 0.96253645615273
2 0.911504858098232
3 0.949981981381513
4 0.969367885953265
5 0.976241155125425
6 0.968936436999995
7 0.973245916120551
8 0.965553141423509
9 0.963369632148022
10 0.982526978112369
11 0.945587960289909
};
\addplot [semithick, blue, mark=*, mark size=1, mark options={solid}, only marks]
table {%
0 0.963071369207911
1 0.960110206449147
2 0.957996195435179
3 0.9742572353794
4 0.968550034435549
5 0.971665256677179
6 0.976645587318713
7 0.974107550204947
8 0.969335425160554
9 0.874566867195917
10 0.950679763258388
11 0.95725460084104
};
\end{axis}

\end{tikzpicture}
\end{subfigure}%
\caption{(Top) Pearson and (Bottom) Spearman correlation between attention and previous layer contribution.}
\label{corrlocal}
\end{figure}

\section{HTA: Local Validation}\label{sec:val}

The ability of attention distributions to provide explanations has been the target of a number of studies \cite{attentionisNotNotexplanation, isattentioninterprable, DeceiveAttention2019}. In particular, \citet{attentionIsNotExplanation} show that attention distributions do not explain the model output and do not correlate well with attribution methods. 
However, if we are exclusively interested in how attention heads behave locally, i.e., without considering their impact on the model's decisions, it is sound to examine attention distributions. The reason for this is that self-attention is the only operation performed in attention heads, and hence, attention distributions precisely represent the information flow within the heads. As a consequence, we can use attention distributions as a reference to validate whether HTA accurately quantifies how information mixes within transformers. 
To verify this, we compare attention distributions to previous layer contribution by computing the correlation between attention maps $\va_{h,i}$ and the contribution $C(\ve_i^{l-1},\vo_{h,j}^l)$ for each head.

A high correlation value would validate HTA as accurately representing the flow of information within transformers.
To calculate the correlation, first, we extract the attention maps for all the heads of BERT for each of the tokens in the examples of our dataset. Then, we pair each attention map to the corresponding contribution. Note, that both attention maps and contributions are distributions that lie in the probability simplex, i.e., all the values are between $0$ and $1$ and their sum is $1$. Next, we calculate Pearson's correlation coefficient for each attention-contribution pair and we aggregate the results into one value per head by computing the mean of the correlation values.

\begin{figure*}[t]
\centering
\begin{subfigure}[t]{1\textwidth}
\includegraphics[width=1\linewidth]{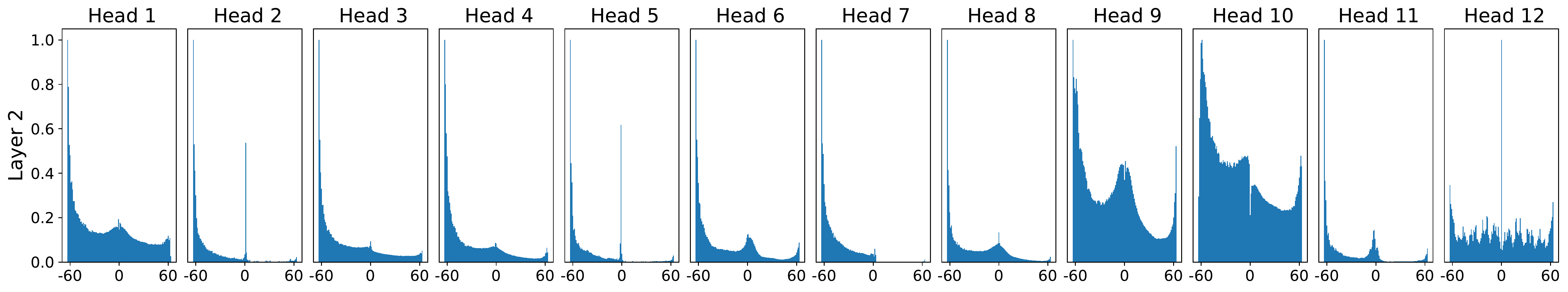}
\end{subfigure}
\begin{subfigure}[t]{1\textwidth}
\includegraphics[width=1\linewidth]{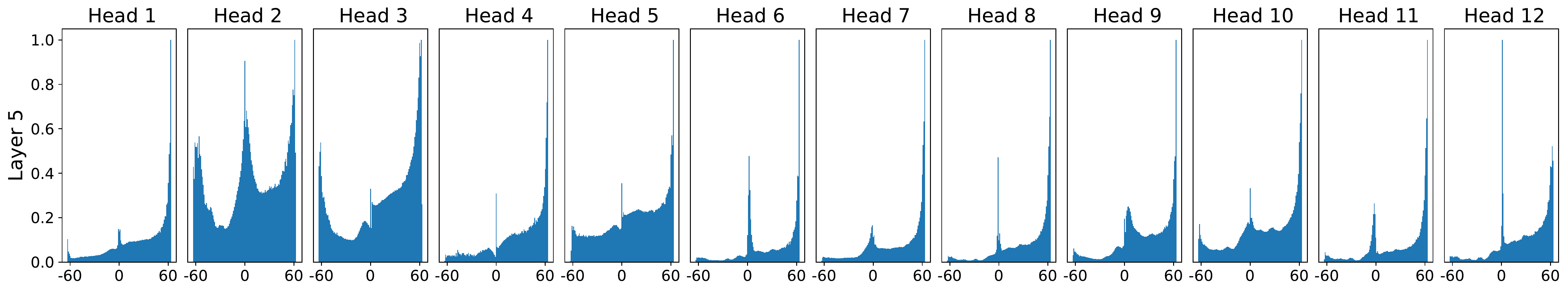}
\end{subfigure}
\begin{subfigure}[t]{1\textwidth}
\includegraphics[width=1\linewidth]{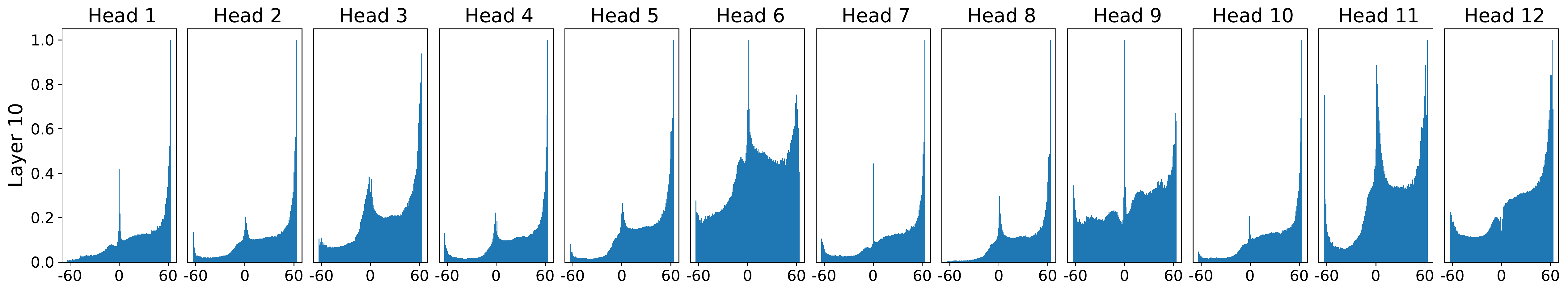}
\end{subfigure}
\caption{Attention histograms for layers 2, 5 and 10 of BERT.  The horizontal axis represents the relative position of the attended tokens with respect to the attending token placed at position 0. Given that the maximum sequence length is 64 the horizontal axis ranges from -63 to 63. The vertical axis is normalized to the maximum value for a better visualization.}
\label{fig:attn}
\end{figure*}

\begin{figure*}[t]
\centering
\begin{subfigure}[t]{1\textwidth}
\includegraphics[width=1\linewidth]{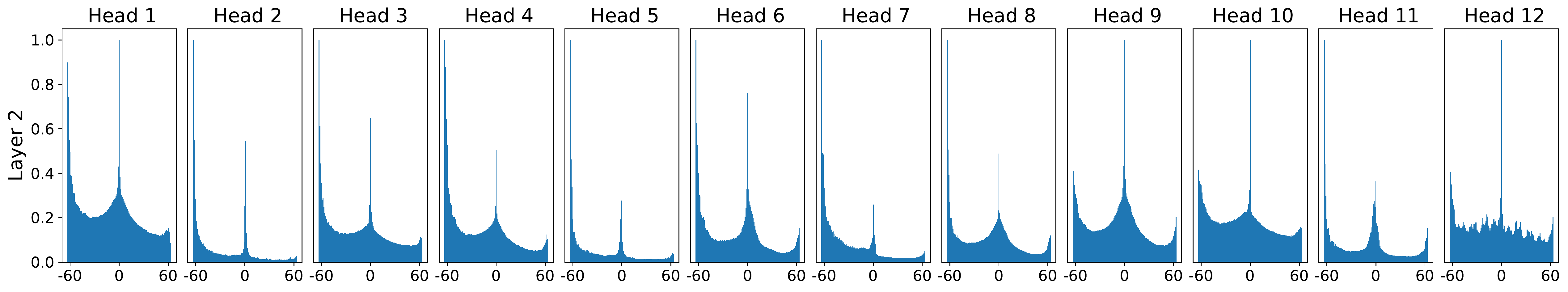}
\end{subfigure}
\begin{subfigure}[t]{1\textwidth}
\includegraphics[width=1\linewidth]{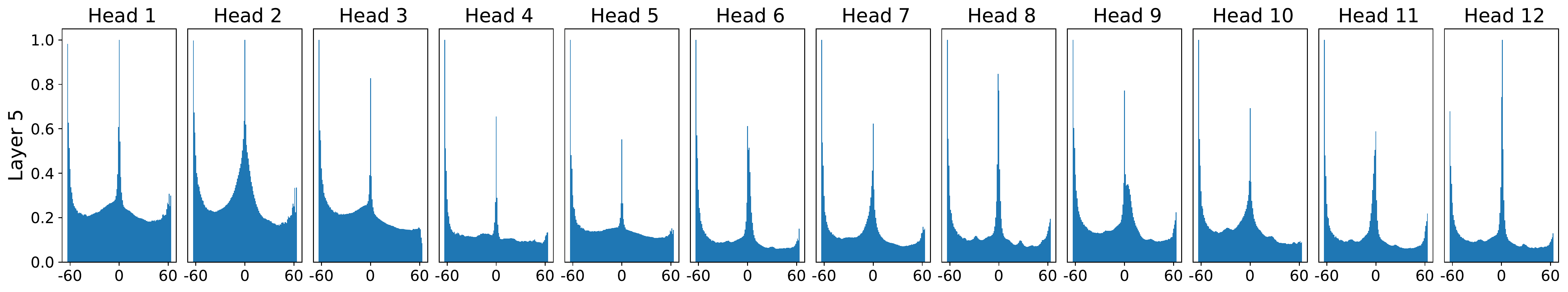}
\end{subfigure}
\begin{subfigure}[t]{1\textwidth}
\includegraphics[width=1\linewidth]{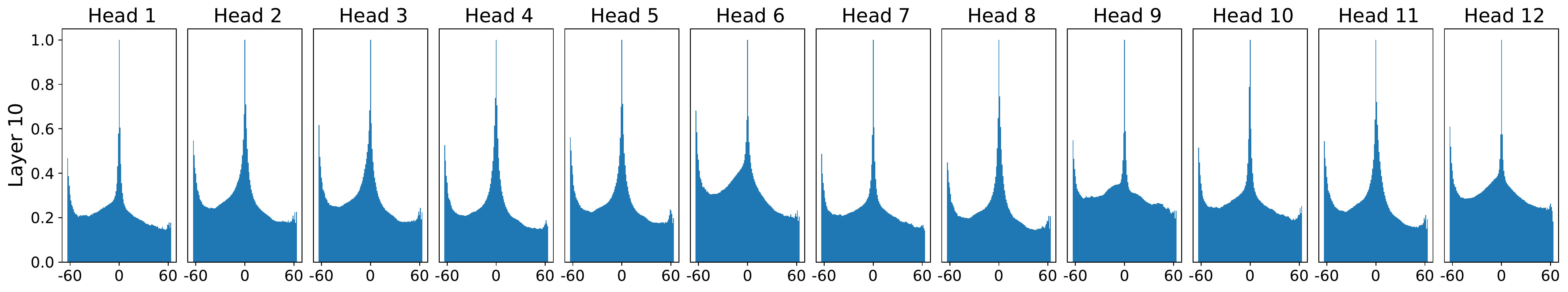}
\end{subfigure}
\caption{Input contribution histograms for layers 2, 5 and 10 of BERT. The horizontal axis represents the relative position of the attended tokens with respect to the attending token placed at position 0. The vertical axis is normalized to the maximum value for a better visualization.}
\label{fig:cont2in}
\end{figure*}

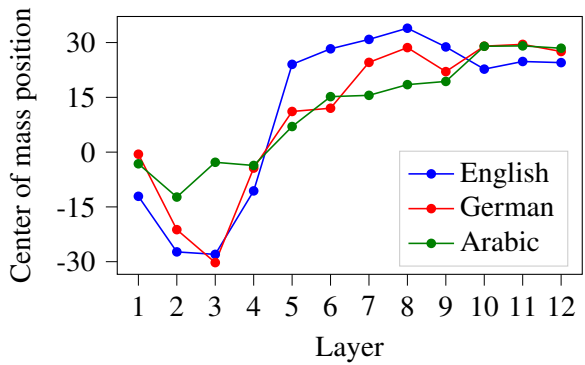
\begin{figure}[t]
\centerfloat
\begin{tikzpicture}

\begin{axis}[
legend cell align={left},
legend style={at={(0.97,0.03)}, anchor=south east, draw=white!80.0!black},
tick align=outside,
tick pos=left,
height = \mainPlotHeight,
width = \linewidth,
x grid style={white!69.01960784313725!black},
xlabel={Layer},
xmin=-0.55, xmax=11.55,
xtick style={color=black},
xtick={0,1,2,3,4,5,6,7,8,9,10,11},
xticklabels={1,2,3,4,5,6,7,8,9,10,11,12},
y grid style={white!69.01960784313725!black},
ylabel={Center of mass position},
ymin=-33.4630084563637, ymax=37.1384940166006,
ytick={-30,-15,0,15,30},
yticklabels={-30,-15,0,15,30},
ytick style={color=black}
]
\addplot [semithick, blue, mark=*, mark size=1.5, mark options={solid}]
table {%
0 -12.097721186175
1 -27.318825516676
2 -27.9966334848606
3 -10.6184486513498
4 24.0314713528331
5 28.2847798452138
6 30.8663103596336
7 33.929334813284
8 28.8035213062002
9 22.7256701683224
10 24.8144369698
11 24.505581635236
};
\addlegendentry{English}
\addplot [semithick, red, mark=*, mark size=1.5, mark options={solid}]
table {%
0 -0.567161333208273
1 -21.2313823789155
2 -30.2538492530472
3 -4.36942063930729
4 11.1298198420788
5 12.0233978553805
6 24.5731115904784
7 28.6162825112768
8 22.0420573517465
9 29.0146548517007
10 29.5033156591305
11 27.5475323734159
};
\addlegendentry{German}
\addplot [semithick, green!50.0!black, mark=*, mark size=1.5, mark options={solid}]
table {%
0 -3.19559974066139
1 -12.305960615007
2 -2.78351068693173
3 -3.65496058877476
4 7.0083509419169
5 15.1926633485686
6 15.5467179424624
7 18.4802618147451
8 19.357029591415
9 28.9838145281447
10 29.088550190467
11 28.4524000788317
};
\addlegendentry{Arabic}
\end{axis}

\end{tikzpicture}
\caption{Mean center of mass for attention histograms per layer for English, German and Arabic.}
\label{cm}
\end{figure}

Figure~\ref{corrlocal} (Top) shows the mean correlation value per head. For all heads except for two, Pearson's correlation coefficient is larger than 0.7. Furthermore, 90\% of the heads show a correlation between attention and Hidden Token Attribution of over 0.85. Similarly, we calculate Spearman's rank correlation coefficient $r$ for each head. The results, displayed in Figure~\ref{corrlocal} (Bottom), show that only four heads have a Spearman's correlation smaller than 0.9, and that 75\% of the heads have a correlation coefficient larger than 0.95. Note that in any case gradient attribution is a local first order approximation and thus introduces a small error that prevents perfect correlation.

These high correlation values empirically demonstrate that 
HTA does indeed represent the flow of information within attention heads with respect to the head inputs.Therefore, to study the inner workings of transformers beyond attention heads one can rely on Hidden Token Attribution and apply it at different points of the model. Now that we have validated HTA,
we can investigate the behavior of the heads in more detail: examining the local patterns revealed by attention, the global patterns revealed by HTA, and the discrepancies between both.

\section{Local Head Analysis}

In this section we take a closer look into the local behaviour of attention heads. Here, local means that we analyze how the intermediate tokens fed into the heads are processed, as opposed to how the model input propagates. To this end, we study attention distributions, but rather than studying each individual example, we aggregate the attention distributions, thus obtaining a general picture of how each head behaves. In particular, we study how much attention is paid to tokens in each relative position with respect to the attending token.

For each head, we extract the attention maps for each token. Then, we define the position of the attending token in the sentence as the origin ($x=0$), thereby generating a histogram where the horizontal axis represents the position of the neighbours and the vertical axis the amount of attention paid to a token. We sum the histograms of all tokens and then normalize the result. To normalize, we divide the value of attention at each position by the number of times that a token is at that relative position; given that the median length of the examples is 36, this normalization ensures that distant positions are not penalized for having fewer occurrences.

Figure~\ref{fig:attn} presents the histograms for the heads in layers 2, 5 and 10, the other layers can be found in Appendix \ref{appatt}. 
From these histograms, a clear pattern is observable. In the first layers, heads tend to aggregate more information from past tokens than from future tokens. In fact, the attention of heads 2, 5 and 7 in Layer 2 to future tokens is negligible. However, this trend quickly reverses with increasing depth, and in later layers the aggregation of future hidden embeddings dominates for most heads. To illustrate this, we calculate the center of mass of the attention histograms per layer by averaging over each head.
The results are shown in Figure~\ref{cm}. To understand whether this behavior is particular to English we perform the same calculations on the German~\cite{GermanBERT} and Arabic~\cite{antoun2020arabert} versions of BERT on 1800 examples of the XNLI dataset~\cite{conneau2018xnli}; note that although Arabic is a right-to-left language, to process the examples the token order is reversed and input to BERT as left-to-right.

The figure shows that BERT follows the same trend regardless of the language, i.e., the models first attend to past and then to future hidden tokens.
This suggests that despite its bidirectional training, BERT tends to handle language like humans, from left to right. This is also inline with the sequential nature of language, i.e., the past context needs to be known to understand the future context.

\section{Global Head Analysis}

Although attention maps are an effective tool to understand the local behavior of attention heads, drawing conclusions that refer to the input words can be misleading. Transformers are complex models that mix information from the entire input sequence at each layer. 
Recent work \cite{Brunner2020On, DeceiveAttention2019} has raised 
concerns about the interpretability of attention maps as representative of global context aggregation. In this section, we look into the individual heads and study what we call global patterns, i.e., aggregation patterns that refer to the model's input. 

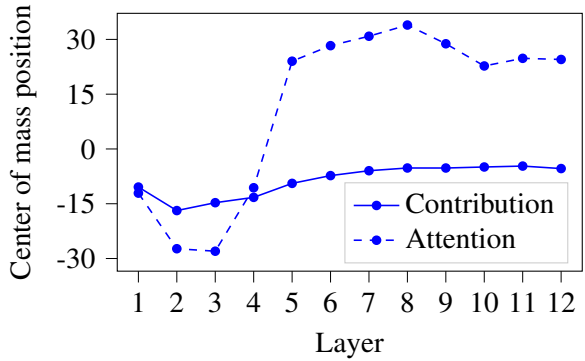
\begin{figure}[t]
\centerfloat
\begin{tikzpicture}

\begin{axis}[
legend cell align={left},
legend style={at={(0.97,0.03)}, anchor=south east, draw=white!80.0!black},
tick align=outside,
tick pos=left,
height = \mainPlotHeight,
width = \linewidth,
x grid style={white!69.01960784313725!black},
xlabel={Layer},
xmin=-0.55, xmax=11.55,
xtick style={color=black},
xtick={0,1,2,3,4,5,6,7,8,9,10,11},
xticklabels={1,2,3,4,5,6,7,8,9,10,11,12},
y grid style={white!69.01960784313725!black},
ylabel={Center of mass position},
ymin=-33.4630084563637, ymax=37.1384940166006,
ytick={-30,-15,0,15,30},
yticklabels={-30,-15,0,15,30},
ytick style={color=black}
]
\addplot [semithick, blue, mark=*, mark size=1.5, mark options={solid}]
table {%
0 -10.4290882380255
1 -16.8710979130467
2 -14.7126206328953
3 -13.2553172699663
4 -9.42239577539838
5 -7.29650610014176
6 -5.95896690944235
7 -5.21205795798401
8 -5.21795160973079
9 -4.9383894037473
10 -4.68637630452656
11 -5.36794450216721
};
\addlegendentry{Contribution}

\addplot [semithick, blue, dashed, mark=*, mark size=1.5, mark options={solid}]
table {%
0 -12.097721186175
1 -27.318825516676
2 -27.9966334848606
3 -10.6184486513498
4 24.0314713528331
5 28.2847798452138
6 30.8663103596336
7 33.929334813284
8 28.8035213062002
9 22.7256701683224
10 24.8144369698
11 24.505581635236
};
\addlegendentry{Attention}
\end{axis}

\end{tikzpicture}
\caption{Mean center of mass for contribution vs. attention histograms per layer for English.}
\label{cm_contr}
\end{figure}

To this end, we follow the same procedure we use in the previous section, but to generate input contribution $C(\ve_i^0, \vo_{h,j}^l)$ histograms instead of attention histograms. In Figure~\ref{fig:cont2in} we show the histograms for layers 2, 5 and 10, i.e., the same layers as in Figure~\ref{fig:attn}. The histograms for the whole model can be found in Appendix~\ref{appcontr}. Furthermore, we calculate the center of mass per layer and compare them to the attention centers of mass in Figure~\ref{cm_contr}. The histograms and the centers of mass show that the global pattern of context aggregation is much more uniform than shown by the attention maps, especially after layer 4. This is intuitive: given that in the first layers the heads are attending mostly to the past context, on average, all the hidden tokens have a larger amount of past context. Therefore, when in later layers the attention shifts to the future hidden tokens, the past context already contained in these tokens balances the contribution, resulting in a uniform pattern of global context aggregation.

The difference in the patterns revealed by this global analysis and the local head analysis from the previous section shows a strong mismatch between attention distributions and global context aggregation in attention heads. In fact, local attention patterns can easily lead to spurious conclusions when used to interpret global context aggregation.
Next, we study this difference quantitatively.

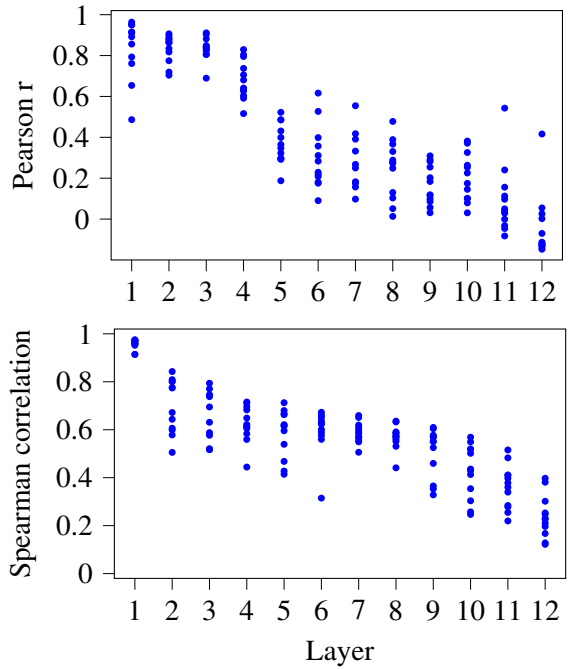
\begin{figure}
\centering
\begin{subfigure}[t]{.47\textwidth}
\centerfloat
\begin{tikzpicture}
\begin{axis}[
tick align=outside,
tick pos=left,
x grid style={white!69.01960784313725!black},
xmin=-0.55, xmax=11.55,
xtick style={color=black},
xtick={0,1,2,3,4,5,6,7,8,9,10,11},
xticklabels={1,2,3,4,5,6,7,8,9,10,11,12},
y grid style={white!69.01960784313725!black},
ylabel={Pearson r},
ymin=-0.2, ymax=1.02,
ytick style={color=black},
ytick={0,0.2,0.4,0.6,0.8,1},
yticklabels={0,0.2,0.4,0.6,0.8,1},
height=0.65\textwidth,
width=\textwidth
]
\addplot [semithick, blue, mark=*, mark size=1, mark options={solid}, only marks]
table {%
0 0.914763753336089
1 0.83447278017983
2 0.911012275671079
3 0.680473151385369
4 0.322812793993617
5 0.398656008281115
6 0.156737566200241
7 0.366594306000914
8 0.0855752703552642
9 0.325817860027186
10 0.0500799807204231
11 -0.113164950908614
};
\addplot [semithick, blue, mark=*, mark size=1, mark options={solid}, only marks]
table {%
0 0.891791980056106
1 0.89795367707273
2 0.830884154334346
3 0.591135979988287
4 0.485928623631706
5 0.208657784284665
6 0.17716432315144
7 0.27325766378159
8 0.0310066737513715
9 0.174778697906311
10 0.0290246434767361
11 -0.117242352332436
};
\addplot [semithick, blue, mark=*, mark size=1, mark options={solid}, only marks]
table {%
0 0.963578297697834
1 0.774704460282613
2 0.805641584076065
3 0.641046594022498
4 0.187567118098392
5 0.0902420544012317
6 0.250088874778924
7 0.279052308459806
8 0.118955756670631
9 0.225489126283188
10 -0.0318771242178039
11 0.00228317410410786
};
\addplot [semithick, blue, mark=*, mark size=1, mark options={solid}, only marks]
table {%
0 0.949513017606226
1 0.863763368595939
2 0.689292705252269
3 0.737185505921973
4 0.347167255854102
5 0.283427641881299
6 0.178316477970771
7 0.129944660532906
8 0.202397308504528
9 0.10140531123169
10 -0.0460711802719372
11 -0.138557798539915
};
\addplot [semithick, blue, mark=*, mark size=1, mark options={solid}, only marks]
table {%
0 0.486384407766893
1 0.886908487670776
2 0.83530014392385
3 0.516169353757704
4 0.292250753530346
5 0.175427904152026
6 0.267585605289379
7 0.477724312191739
8 0.0991328790178775
9 0.145626893630857
10 0.0377448099330655
11 -0.0701854211006428
};
\addplot [semithick, blue, mark=*, mark size=1, mark options={solid}, only marks]
table {%
0 0.856122789219537
1 0.817681588624152
2 0.822646011358214
3 0.803742158853663
4 0.522423678977136
5 0.180661001283055
6 0.390919878040526
7 0.28834061296854
8 0.116391132948328
9 0.371192534528027
10 -0.0825881993424873
11 0.0259950432149899
};
\addplot [semithick, blue, mark=*, mark size=1, mark options={solid}, only marks]
table {%
0 0.653916609361279
1 0.906204216586191
2 0.881388007970676
3 0.626691907325525
4 0.399092671125045
5 0.229118916173002
6 0.33252551624469
7 0.0510675510506717
8 0.0565281985860498
9 0.101098716898559
10 0.0969495617868786
11 -0.147643237692631
};
\addplot [semithick, blue, mark=*, mark size=1, mark options={solid}, only marks]
table {%
0 0.760863733938521
1 0.880753753648266
2 0.805181130066833
3 0.633983483655977
4 0.364670804886707
5 0.357242637189975
6 0.180363344521882
7 0.0133004025235557
8 0.25405455964592
9 0.263523210051608
10 -0.000774058286906137
11 -0.12869589695569
};
\addplot [semithick, blue, mark=*, mark size=1, mark options={solid}, only marks]
table {%
0 0.911458167174228
1 0.719167912342388
2 0.842471548668816
3 0.604279464245397
4 0.295362750369558
5 0.311025099408222
6 0.0977161076869703
7 0.330757175277
8 0.183891233389631
9 0.381726417205102
10 0.240216085083256
11 0.416236597656337
};
\addplot [semithick, blue, mark=*, mark size=1, mark options={solid}, only marks]
table {%
0 0.793178007582768
1 0.704038872384086
2 0.903340435292561
3 0.829314371375638
4 0.298321977383277
5 0.616330462333428
6 0.417497545977218
7 0.248768856191257
8 0.309444135490214
9 0.0305875593005652
10 0.15602880335019
11 0.0547301271448057
};
\addplot [semithick, blue, mark=*, mark size=1, mark options={solid}, only marks]
table {%
0 0.950848601272698
1 0.863893939937558
2 0.841421594448203
3 0.705746150559029
4 0.485315426484223
5 0.526665232015108
6 0.182386190418552
7 0.388785465386394
8 0.289946152455683
9 0.253325577039762
10 0.542833407692665
11 -0.125897137682825
};
\addplot [semithick, blue, mark=*, mark size=1, mark options={solid}, only marks]
table {%
0 0.914066874770735
1 0.87353786441198
2 0.848937854642612
3 0.795290972067281
4 0.430996114725697
5 0.218843930316334
6 0.554572298335898
7 0.103265431136952
8 0.282125559650441
9 0.0793407332259496
10 0.112812183204592
11 -0.124627015786893
};
\end{axis}

\end{tikzpicture}
\end{subfigure}
\hfill
\begin{subfigure}[t]{.47\textwidth}
\centerfloat
\begin{tikzpicture}

\begin{axis}[
tick align=outside,
tick pos=left,
x grid style={white!69.01960784313725!black},
xlabel={Layer},
xmin=-0.55, xmax=11.55,
xtick style={color=black},
xtick={0,1,2,3,4,5,6,7,8,9,10,11},
xticklabels={1,2,3,4,5,6,7,8,9,10,11,12},
y grid style={white!69.01960784313725!black},
ylabel={Spearman correlation},
ymin=-0.02, ymax=1.02,
ytick style={color=black},
ytick={0,0.2,0.4,0.6,0.8,1},
yticklabels={0,0.2,0.4,0.6,0.8,1},
height=0.65\textwidth,
width=\textwidth
]
\addplot [semithick, blue, mark=*, mark size=1, mark options={solid}, only marks]
table {%
0 0.958729802611312
1 0.774053346776016
2 0.520927939939614
3 0.444272746968773
4 0.539192962633458
5 0.660079456012838
6 0.60079905178182
7 0.573958458012723
8 0.551269225160722
9 0.519513189720443
10 0.219581703408481
11 0.230064266700868
};
\addplot [semithick, blue, mark=*, mark size=1, mark options={solid}, only marks]
table {%
0 0.967840870290622
1 0.578357328972624
2 0.694919637789383
3 0.69605886059797
4 0.595688394760118
5 0.625009994894545
6 0.549478289778821
7 0.587230742315833
8 0.353012530115795
9 0.430182864511108
10 0.515586405491973
11 0.167167042259726
};
\addplot [semithick, blue, mark=*, mark size=1, mark options={solid}, only marks]
table {%
0 0.97511974605117
1 0.671792147695919
2 0.587898153005864
3 0.559470446934995
4 0.414282639010131
5 0.314777307350712
6 0.619447507320367
7 0.560940739623409
8 0.328363491734204
9 0.435959129560425
10 0.27835684070864
11 0.227494462592599
};
\addplot [semithick, blue, mark=*, mark size=1, mark options={solid}, only marks]
table {%
0 0.972911287669394
1 0.801812594339646
2 0.517171640209447
3 0.715685200630072
4 0.427730410572391
5 0.652386045060021
6 0.584817878479152
7 0.571677619833434
8 0.525374361929011
9 0.413114082439208
10 0.254930312339851
11 0.210933851031135
};
\addplot [semithick, blue, mark=*, mark size=1, mark options={solid}, only marks]
table {%
0 0.914278283489141
1 0.599575566346056
2 0.737953336415112
3 0.605917891628754
4 0.467960487255222
5 0.559834218927443
6 0.659085116978119
7 0.530779089507721
8 0.550725097591381
9 0.501672411645791
10 0.378017086305575
11 0.253017320207852
};
\addplot [semithick, blue, mark=*, mark size=1, mark options={solid}, only marks]
table {%
0 0.952225668664546
1 0.800244953258737
2 0.770395135213932
3 0.583489971575874
4 0.67992064222192
5 0.576109171819
6 0.651006269001766
7 0.6326053240252
8 0.576949313502923
9 0.51913184883449
10 0.360216378366226
11 0.381148647244802
};
\addplot [semithick, blue, mark=*, mark size=1, mark options={solid}, only marks]
table {%
0 0.958685620492179
1 0.505459279518742
2 0.576834184598839
3 0.616542449963241
4 0.615050262772237
5 0.601035584539358
6 0.570595745029596
7 0.583131764798584
8 0.459459530929644
9 0.24670480566484
10 0.482833491333535
11 0.122102938993139
};
\addplot [semithick, blue, mark=*, mark size=1, mark options={solid}, only marks]
table {%
0 0.958829588002241
1 0.842749451290677
2 0.743556003095446
3 0.609868242147169
4 0.666707368388391
5 0.626237374861874
6 0.556546036142397
7 0.441120891297493
8 0.568958795197164
9 0.568847951617852
10 0.283975712369448
11 0.397659487326882
};
\addplot [semithick, blue, mark=*, mark size=1, mark options={solid}, only marks]
table {%
0 0.96926794720037
1 0.775738346577282
2 0.745845005837186
3 0.648660427149478
4 0.66268746288646
5 0.672836547313672
6 0.505958028567895
7 0.551633180707216
8 0.365571337921158
9 0.354037643364915
10 0.339728290554482
11 0.248042526199652
};
\addplot [semithick, blue, mark=*, mark size=1, mark options={solid}, only marks]
table {%
0 0.913850306461375
1 0.643539027325886
2 0.517194835013986
3 0.621759621703028
4 0.618922373650398
5 0.589386037982062
6 0.608501457454936
7 0.590944450065291
8 0.609496370812926
9 0.257404165128241
10 0.397305628616147
11 0.128159539996361
};
\addplot [semithick, blue, mark=*, mark size=1, mark options={solid}, only marks]
table {%
0 0.97336395567679
1 0.808877368487041
2 0.630986656708766
3 0.710793041175108
4 0.712769896997367
5 0.638081709781316
6 0.589676497835313
7 0.635661232736607
8 0.604365224388664
9 0.549263988037425
10 0.411547968289794
11 0.196626434220799
};
\addplot [semithick, blue, mark=*, mark size=1, mark options={solid}, only marks]
table {%
0 0.963023411331783
1 0.606102761496575
2 0.793835951675577
3 0.682562988177106
4 0.621034521362762
5 0.624196085002465
6 0.569798870964144
7 0.569242342945677
8 0.57107027034912
9 0.303635756437292
10 0.409735236675423
11 0.30163211241635
};
\end{axis}

\end{tikzpicture}
\end{subfigure}%
\caption{(Top) Pearson and (Bottom) Spearman correlation between attention and input contribution.}
\label{fig:corr2in}
\end{figure}

\section{Local Attention vs. Global Attribution}\label{sec:attnvsattr}

To quantify the discrepancy between attention distributions and input contribution, i.e., local and global patterns of context aggregation, we calculate the correlation between attention maps and input contribution $C(\ve_i^0, \vo_{h,j}^l)$. We follow the same methodology as in Section~\ref{sec:val} and report Pearson's and Spearman's correlation coefficient in Figure~\ref{fig:corr2in}. In line with the mismatch between attention and contribution histograms (Figures~\ref{fig:attn} and \ref{fig:cont2in}), we observe how the correlation between attention and input contribution quickly decreases in deeper layers. Particularly, after only four layers Pearson's correlation coefficient for most heads is smaller than $0.5$ and in the last four layers the median head correlation value is smaller than $0.25$. Furthermore, Spearman's correlation follows a very similar trend, with the median head correlation value falling under $0.7$ already at layer 3, and under $0.25$ at the last layer.

The results from this section point at the importance of information mixing: attention maps show how the heads behave locally, i.e., how they aggregate context, but not \emph{what} context is in fact aggregated. Knowing how the heads behave locally can give us a better understanding of transformer models that could be leveraged to further improve the performance of these models~\cite{Wu2020Lite}. However, attention maps
are misleading when drawing conclusions about what input words are being aggregated into the contextual embeddings.

\subsection{Specific examples}

\begin{figure*}[t]
\centering
\includegraphics[width=1\linewidth]{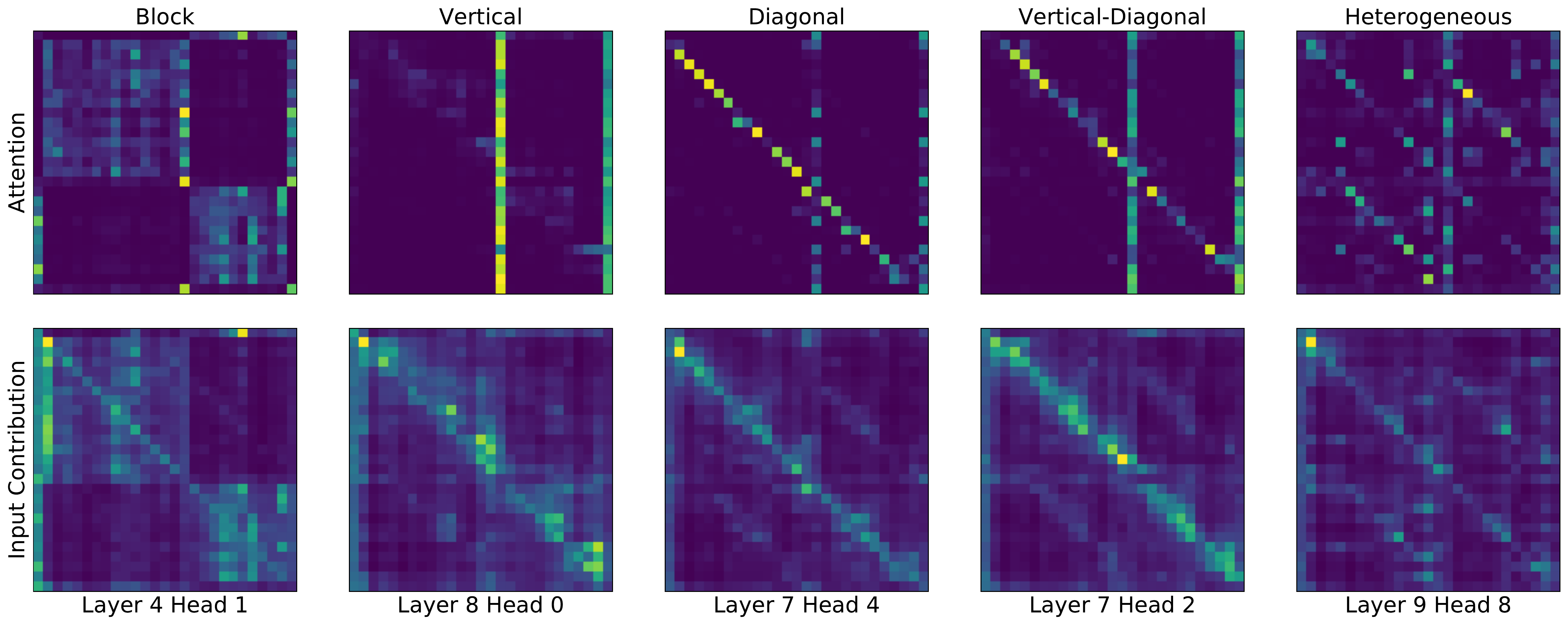}
\caption{Comparison of the head patterns revealed by attention distributions (upper row) and input contribution (lower row).}
\label{fig:headtypes}
\end{figure*}

The histograms studied in the previous sections give us a high level picture of what is happening inside the model. However, we averaged across examples with different sequence length and with different token types in different positions. 
To gain a more detailed understanding of the model's behaviour, we now look into specific input sequences randomly selected from our dataset. 

\citet{darksecrets} study attention maps generated by BERT for many different examples and divide the attention patterns into five types: vertical, diagonal, vertical-diagonal, block and heterogeneous. When looking at the attention maps, we observe the same five attention patterns. Nevertheless,
to understand what input information these heads are actually aggregating, we need to look at the contribution from the input tokens. 

In Figure~\ref{fig:headtypes}, we compare the five patterns observed by \citet{darksecrets} with the corresponding patterns revealed by Hidden Token Attribution with respect to the input, $C(\ve_i^0, \vo_{h,j}^l)$. A comparison for all heads is available in Appendix~\ref{appcompa}. Remarkably, heads with the vertical pattern pay most attention to the SEP and CLS tokens. Nevertheless, the input contribution reveals that SEP tokens are used by the model to store general context, and by extracting information from the SEP token at intermediate layers, the model is in fact aggregating global context. Hence, with respect to the input, heads with vertical, diagonal and vertical-diagonal patterns have a similar behavior to heterogeneous heads. However, tokens around the diagonal tend to contribute the most given the prevalent aggregation of local context. 

On the other hand, as shown in the first column of Figure~\ref{fig:headtypes}, we observe that the block patterns prevail when we apply Hidden Token Attribution to the input. It is noteworthy that, while vertical and diagonal patterns fade away, the block pattern still remains visible. The fact that attending to tokens inside a block results in aggregation of context from within that block implies that up to that point, the context was mainly aggregated from within the blocks separated by SEP. We do not observe block patterns in the contribution maps for layers deeper than layer 4, which suggests that the first layers aggregate context within blocks and later layers aggregate context in a more global manner. 

\section{Conclusion}\label{discussion}

We provide justification for using HTA to study information flow within transformers.
By studying attention distributions of BERT we uncover an interesting pattern: In earlier layers, attention heads attend mostly to earlier tokens, whereas this trend quickly reverses with increasing depth. This is surprising, since BERT is trained using bi-directional language modeling and it suggests that like humans, BERT understands language from left to right. 

A problem with local attention patterns is that they do not reveal how the attention heads process the information contained in the input tokens. 
We thus use Hidden Token Attribution to compute per-head attribution distributions over the input words. Our results show that the mismatch between attention and attribution distributions increases with depth. This confirms the importance of accounting for information mixing when analyzing attention heads with respect to the input tokens. Finally, we show how five different attention head patterns differ from their token attribution equivalents. Our method and results are complementary to those in~\cite{abnar2020zuidema}, we believe a combination of attention flow and HTA may provide new interesting insights. 

In this work, we aim to set a clear border that distinguishes between local and global aggregation in transformer models. This distinction is important when trying to interpret the behavior of these models, and we hope that it will help future studies in their analyses. Furthermore, our findings add new insights to the growing field of research on explaining transformers. This research, in turn, can help in guiding design decisions leading to further improvements of natural language processing architectures.

\section*{Acknowledgements}

We would like to thank our colleagues Oliver Richter, Lukas Faber and B\'eni Egressy for the insightful discussions and helpful feedback on preliminary versions of this work.

\bibliography{eacl2021}
\bibliographystyle{acl_natbib}

\newpage

\appendix

\onecolumn
\clearpage
\section{Attention Histograms English}\label{appatt}

\begin{figure*}[h!]
\centering
\begin{subfigure}[t]{1\textwidth}
\includegraphics[width=1\linewidth]{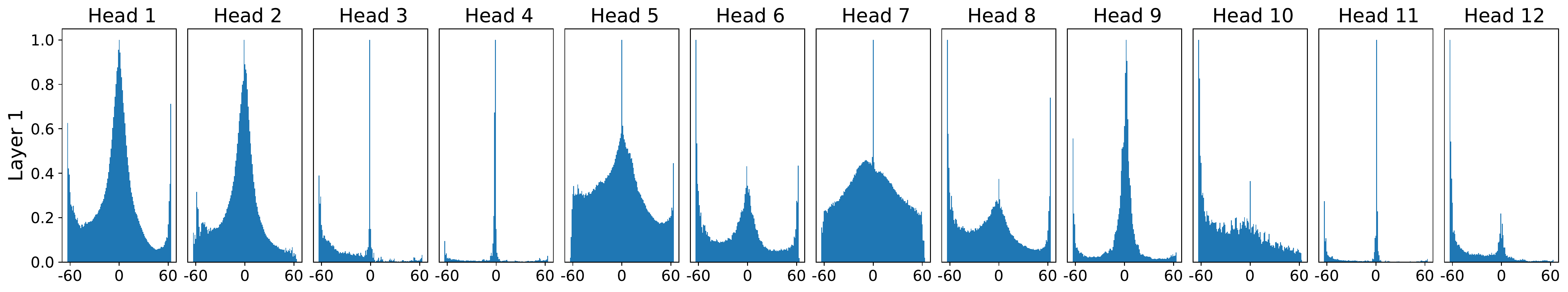}
\end{subfigure}
\begin{subfigure}[t]{1\textwidth}
\includegraphics[width=1\linewidth]{./figures/attention_notft/attn_per_token_l1_upto64.pdf}
\end{subfigure}
\begin{subfigure}[t]{1\textwidth}
\includegraphics[width=1\linewidth]{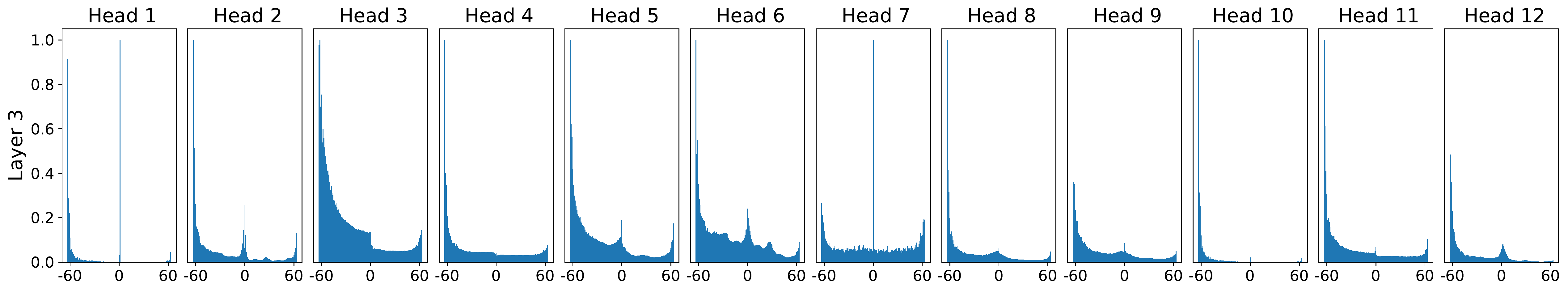}
\end{subfigure}
\begin{subfigure}[t]{1\textwidth}
\includegraphics[width=1\linewidth]{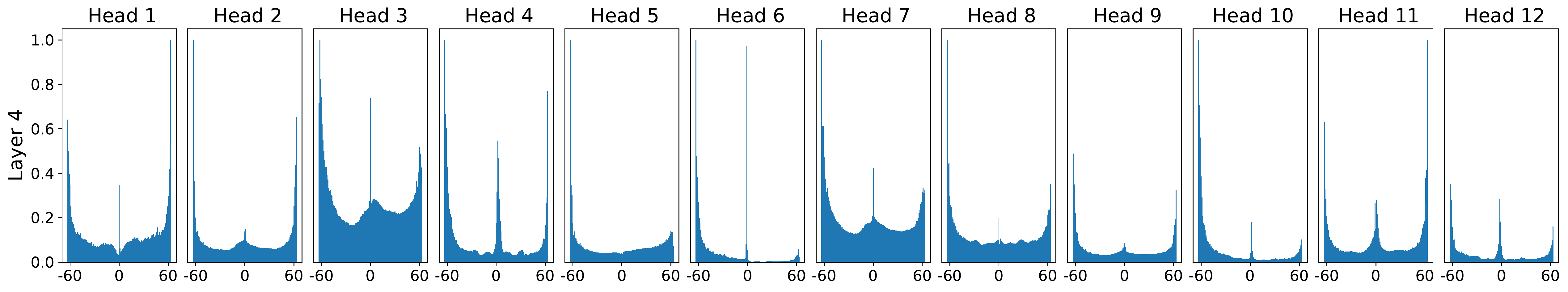}
\end{subfigure}
\begin{subfigure}[t]{1\textwidth}
\includegraphics[width=1\linewidth]{./figures/attention_notft/attn_per_token_l4_upto64.pdf}
\end{subfigure}
\begin{subfigure}[t]{1\textwidth}
\includegraphics[width=1\linewidth]{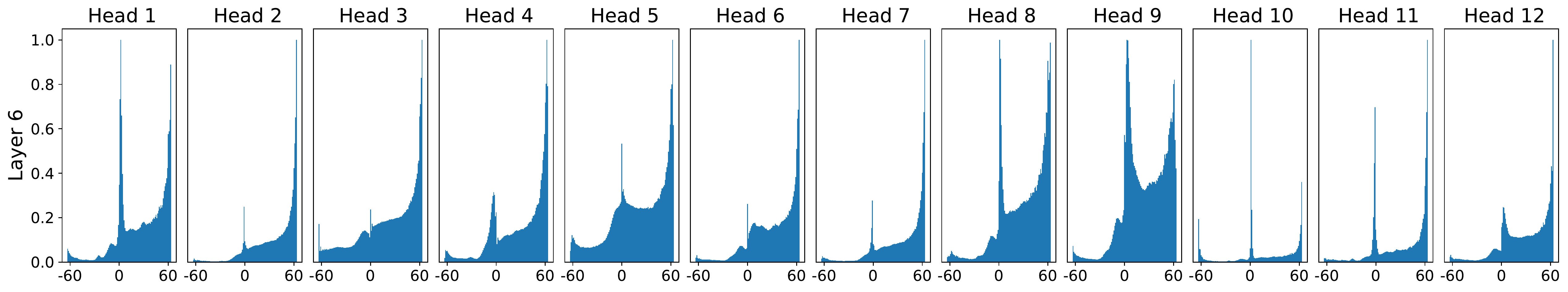}
\end{subfigure}
\caption{Attention histograms for layers 1 to 6}
\label{fig:atthistapp1}
\end{figure*}
\clearpage

\begin{figure*}
\centering
\begin{subfigure}[t]{1\textwidth}
\includegraphics[width=1\linewidth]{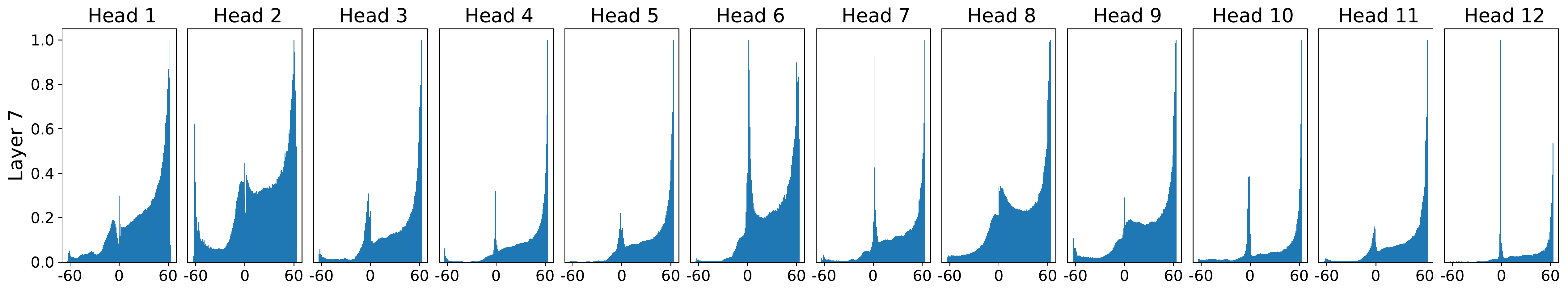}
\end{subfigure}
\begin{subfigure}[t]{1\textwidth}
\includegraphics[width=1\linewidth]{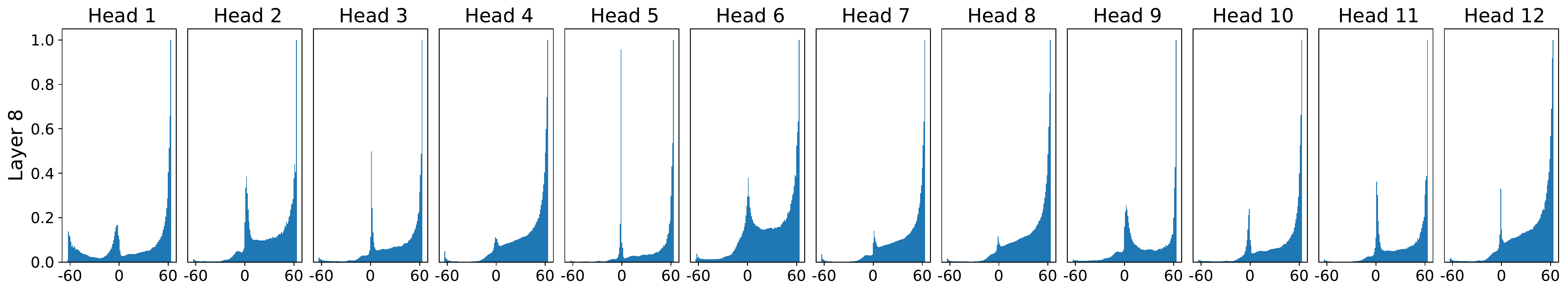}
\end{subfigure}
\begin{subfigure}[t]{1\textwidth}
\includegraphics[width=1\linewidth]{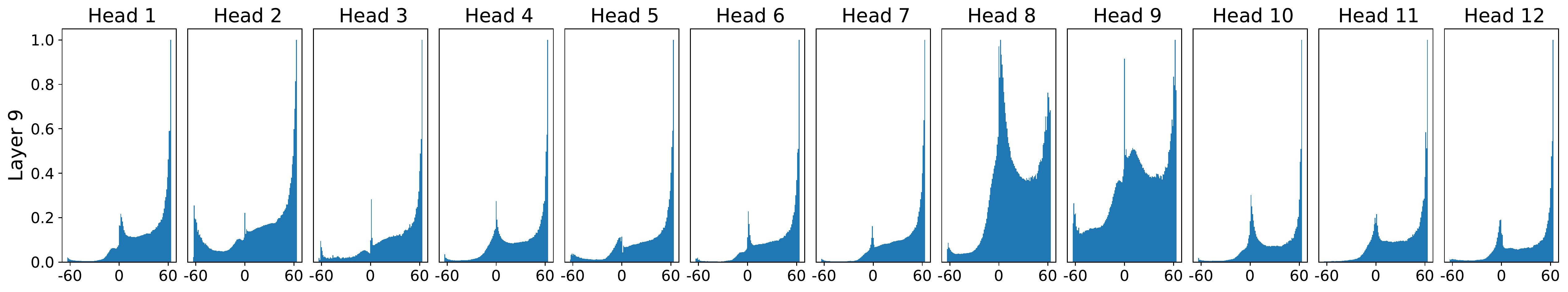}
\end{subfigure}
\begin{subfigure}[t]{1\textwidth}
\includegraphics[width=1\linewidth]{./figures/attention_notft/attn_per_token_l9_upto64.pdf}
\end{subfigure}
\begin{subfigure}[t]{1\textwidth}
\includegraphics[width=1\linewidth]{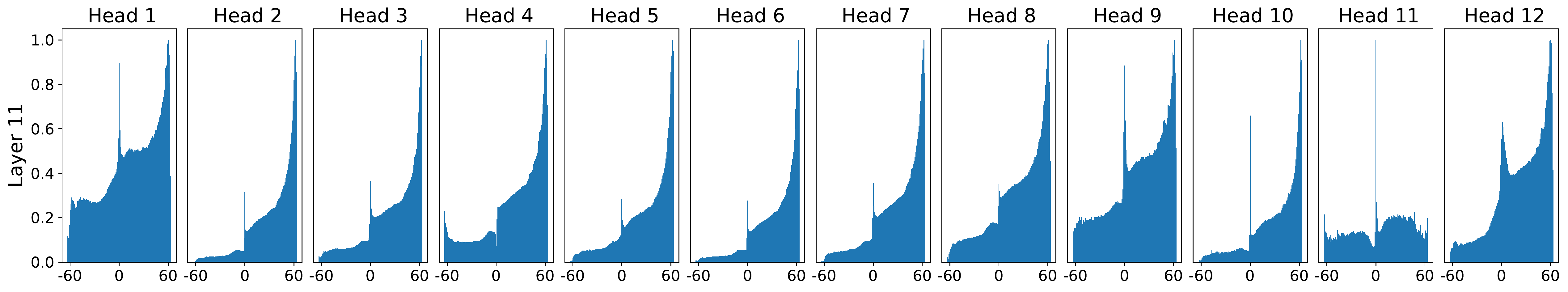}
\end{subfigure}
\begin{subfigure}[t]{1\textwidth}
\includegraphics[width=1\linewidth]{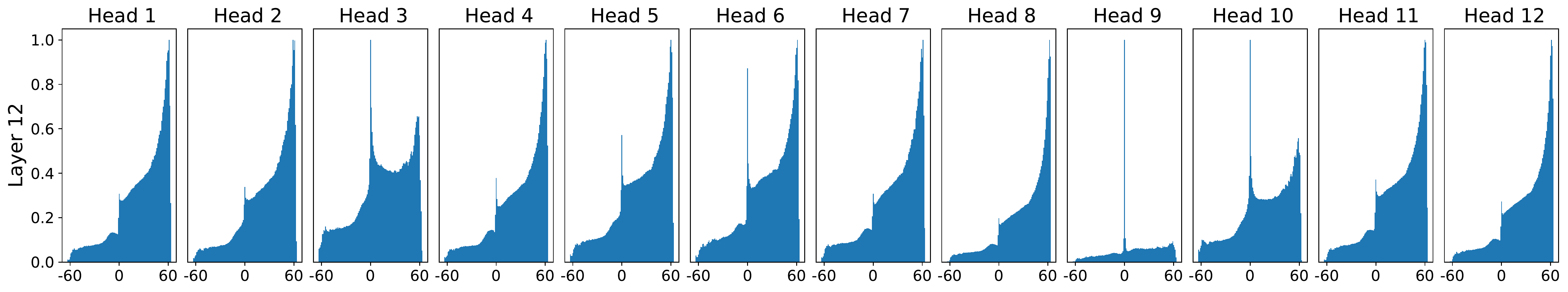}
\end{subfigure}
\caption{Attention histograms for layers 7 to 12}
\label{fig:atthistapp2}
\end{figure*}

\onecolumn
\clearpage
\section{Input Contribution Histograms}\label{appcontr}

\begin{figure*}[h!]
\centering
\begin{subfigure}[t]{1\textwidth}
\includegraphics[width=1\linewidth]{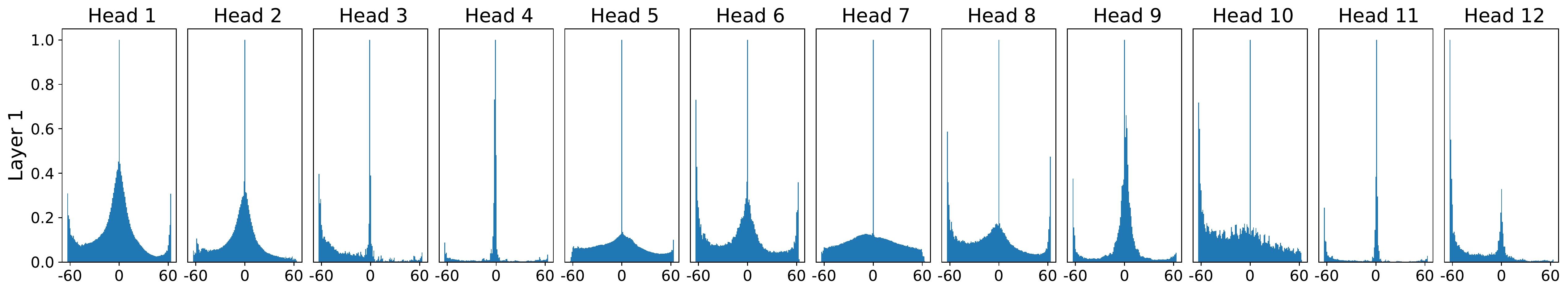}
\end{subfigure}
\begin{subfigure}[t]{1\textwidth}
\includegraphics[width=1\linewidth]{./figures/context_head2in/context_per_token_head2in_l1_upto64.pdf}
\end{subfigure}
\begin{subfigure}[t]{1\textwidth}
\includegraphics[width=1\linewidth]{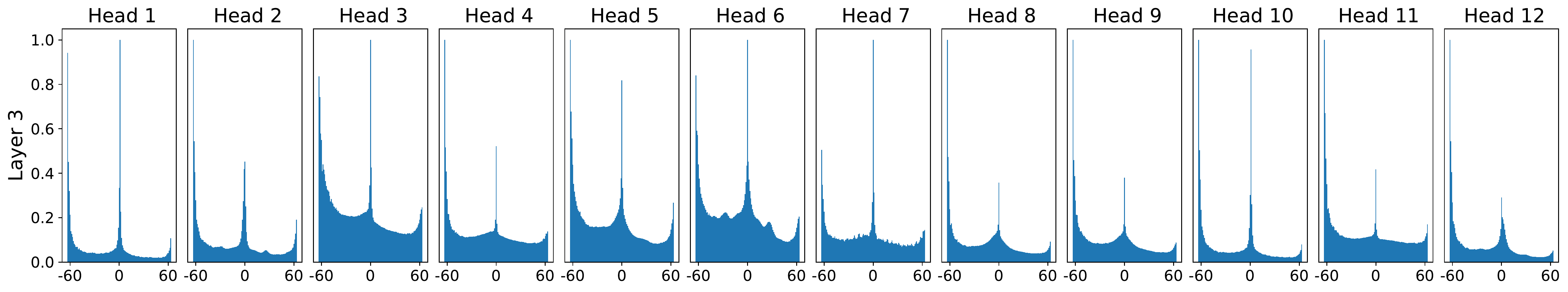}
\end{subfigure}
\begin{subfigure}[t]{1\textwidth}
\includegraphics[width=1\linewidth]{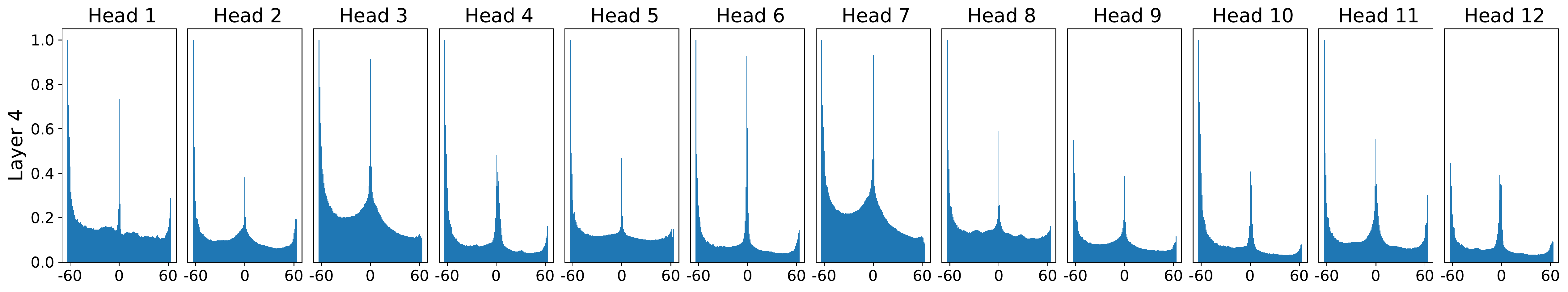}
\end{subfigure}
\begin{subfigure}[t]{1\textwidth}
\includegraphics[width=1\linewidth]{./figures/context_head2in/context_per_token_head2in_l4_upto64.pdf}
\end{subfigure}
\begin{subfigure}[t]{1\textwidth}
\includegraphics[width=1\linewidth]{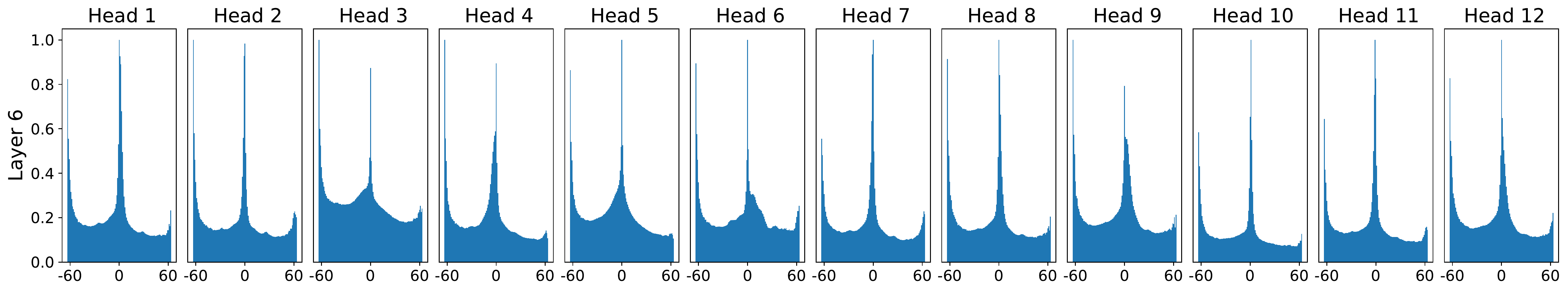}
\end{subfigure}
\caption{Input contribution histograms for layers 1 to 6}
\label{fig:incontrhist1}
\end{figure*}
\clearpage

\begin{figure*}
\centering
\begin{subfigure}[t]{1\textwidth}
\includegraphics[width=1\linewidth]{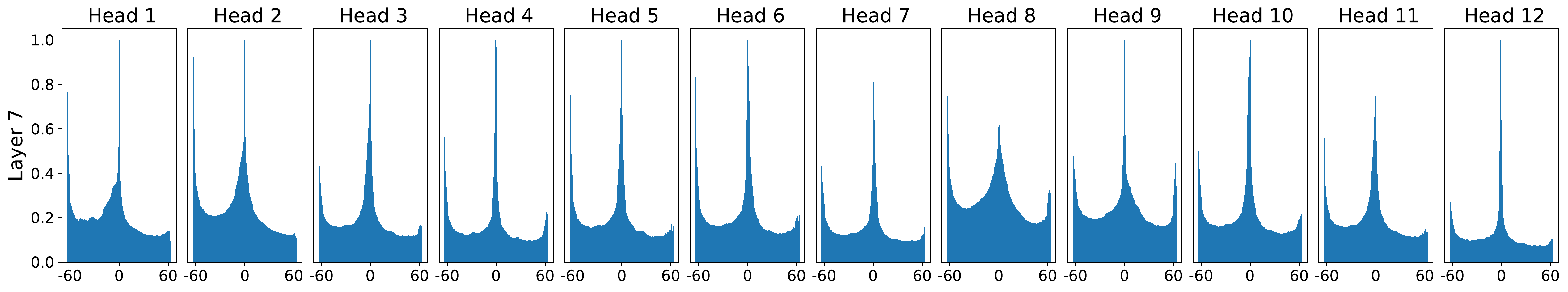}
\end{subfigure}
\begin{subfigure}[t]{1\textwidth}
\includegraphics[width=1\linewidth]{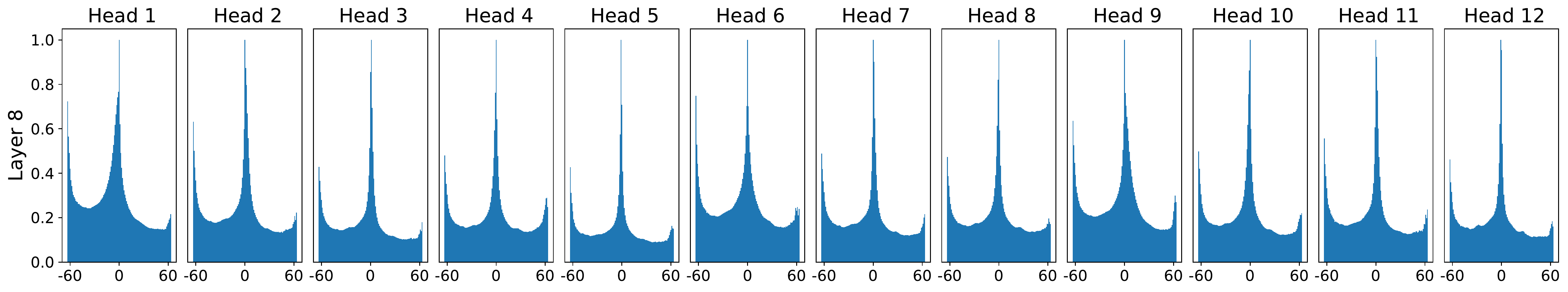}
\end{subfigure}
\begin{subfigure}[t]{1\textwidth}
\includegraphics[width=1\linewidth]{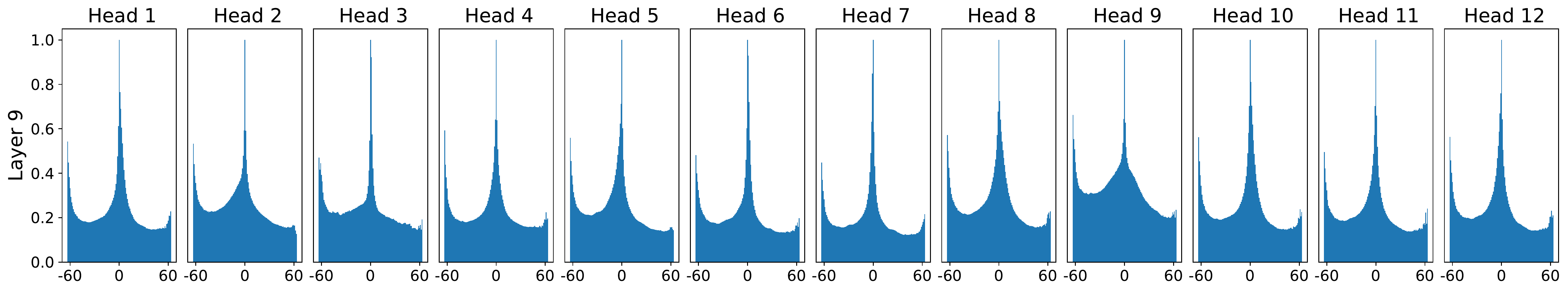}
\end{subfigure}
\begin{subfigure}[t]{1\textwidth}
\includegraphics[width=1\linewidth]{./figures/context_head2in/context_per_token_head2in_l9_upto64.pdf}
\end{subfigure}
\begin{subfigure}[t]{1\textwidth}
\includegraphics[width=1\linewidth]{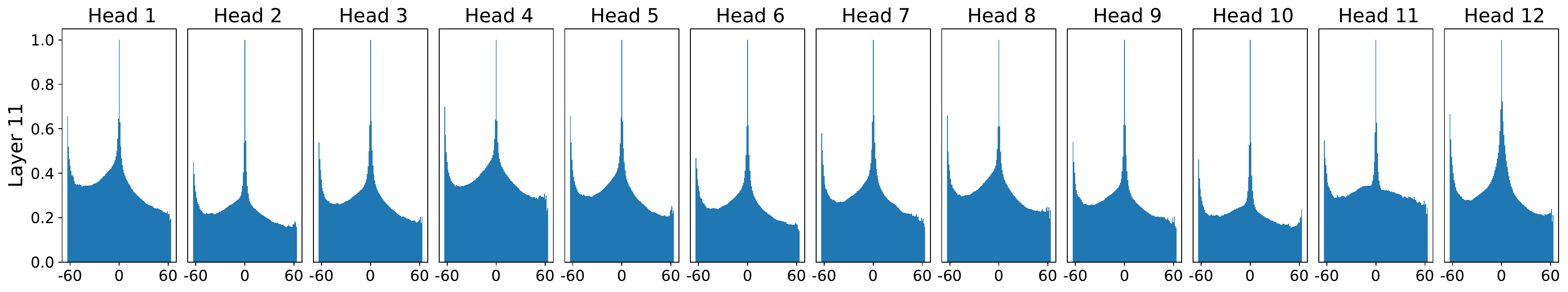}
\end{subfigure}
\begin{subfigure}[t]{1\textwidth}
\includegraphics[width=1\linewidth]{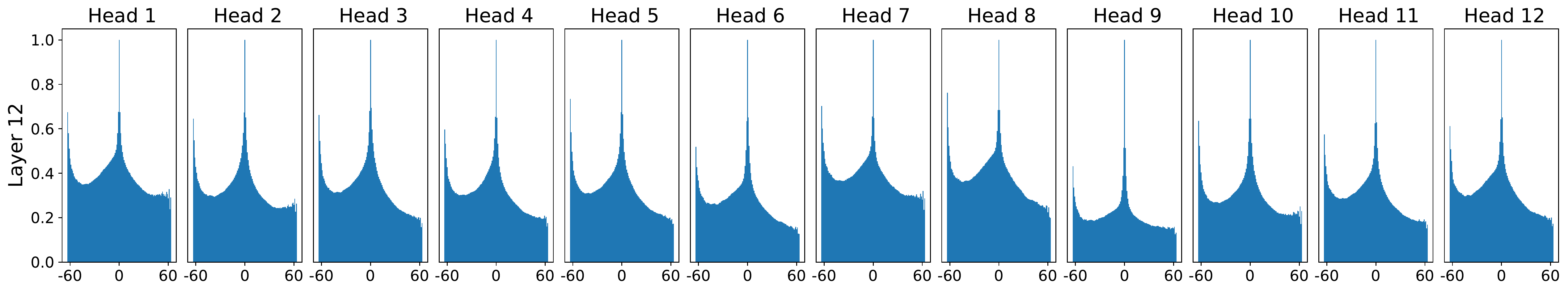}
\end{subfigure}
\caption{Input contribution histograms for layers 7 to 12}
\label{fig:incontrhist2}
\end{figure*}

\onecolumn

\section{Comparison of Local vs. Global Head Patterns}\label{appcompa}
\begin{figure*}[h!]
\centering
\includegraphics[width=0.85\linewidth]{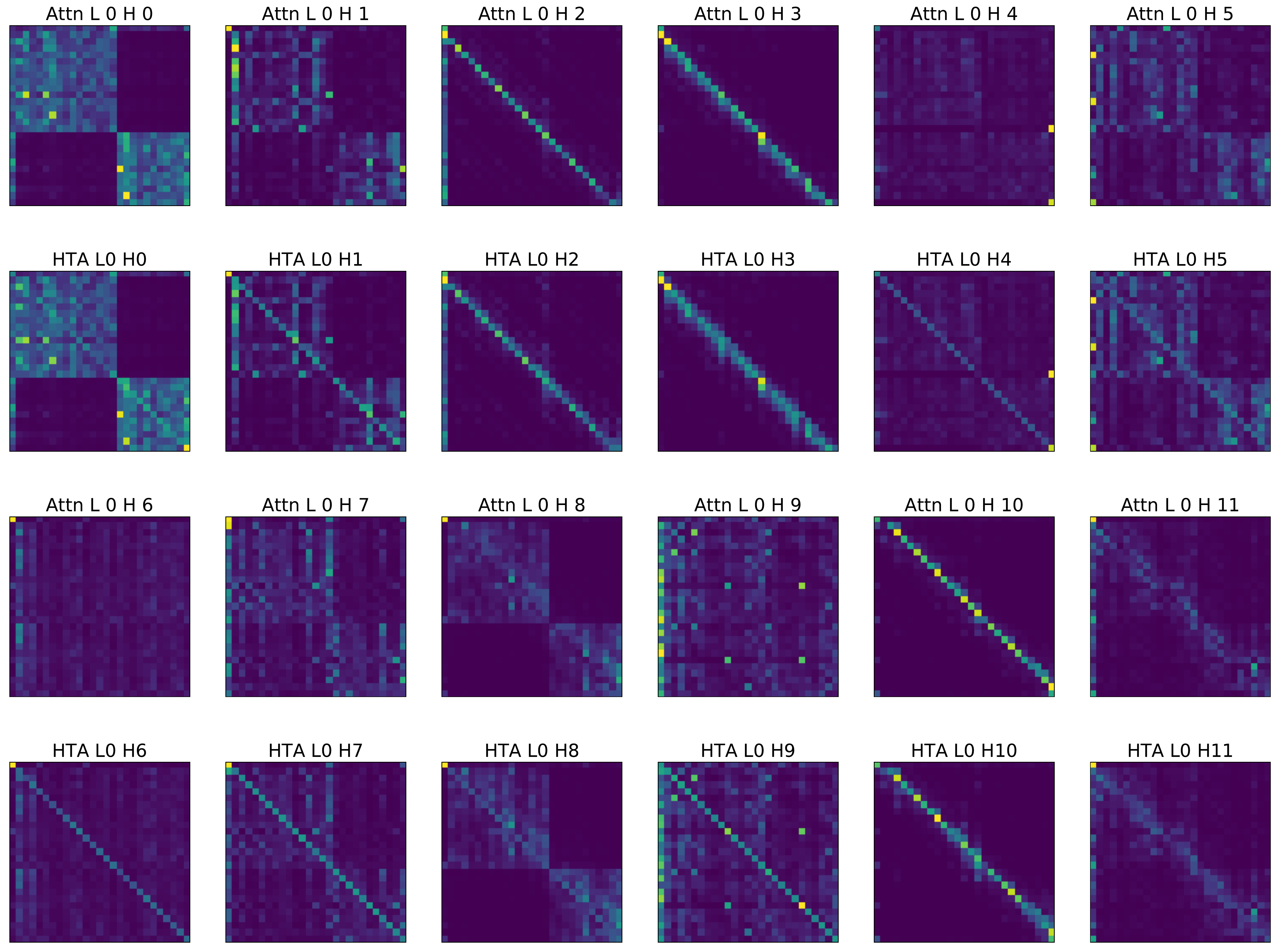}
\caption{Layer 1: rows 1 and 3 represent attention maps, rows 2 and 4 show the corresponding contribution maps.}
\end{figure*}
\begin{figure*}[h!]
\centering
\includegraphics[width=0.85\linewidth]{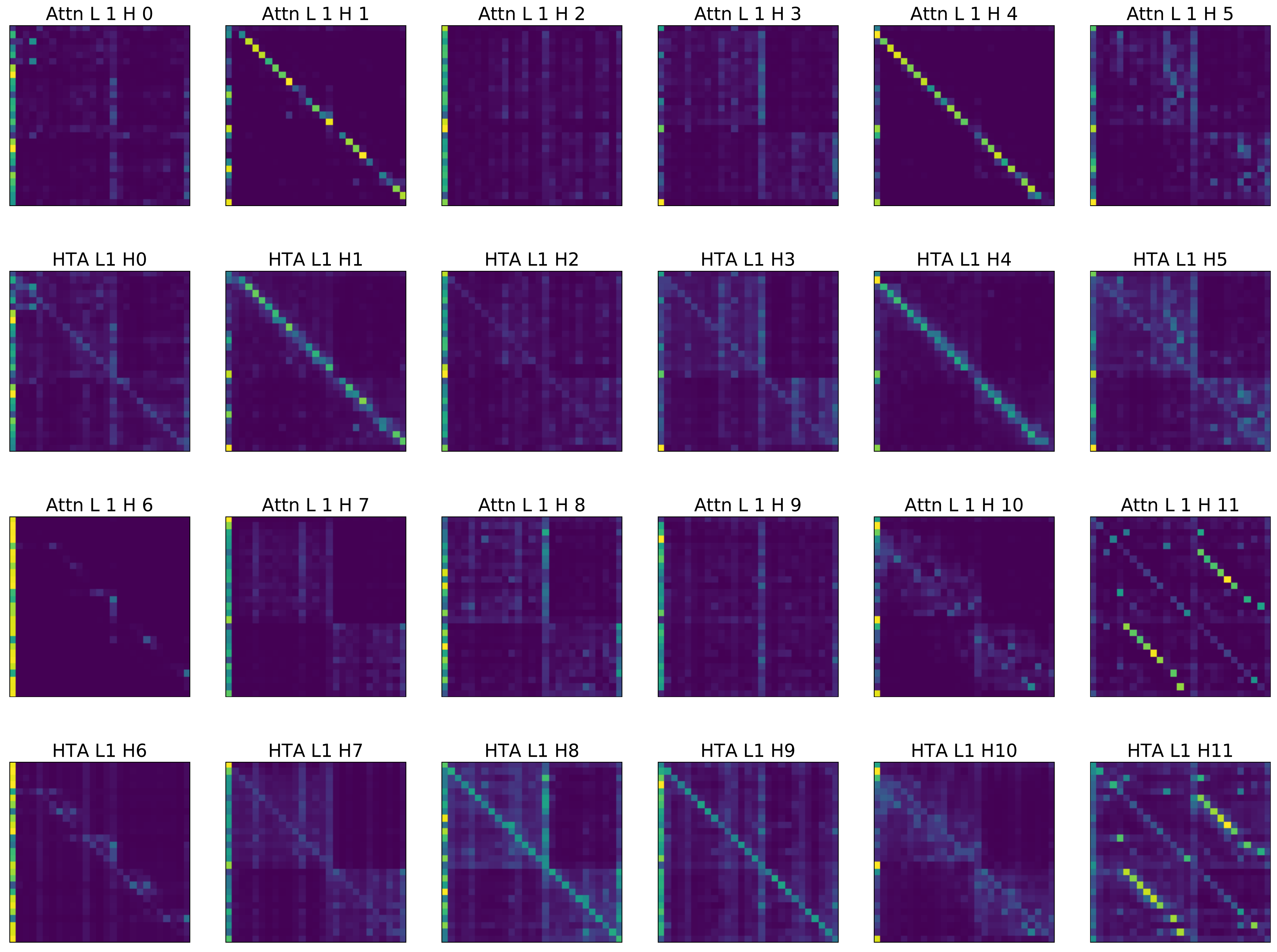}
\caption{Layer 2: rows 1 and 3 represent attention maps, rows 2 and 4 show the corresponding contribution maps.}
\end{figure*}

\begin{figure*}
\centering
\includegraphics[width=0.85\linewidth]{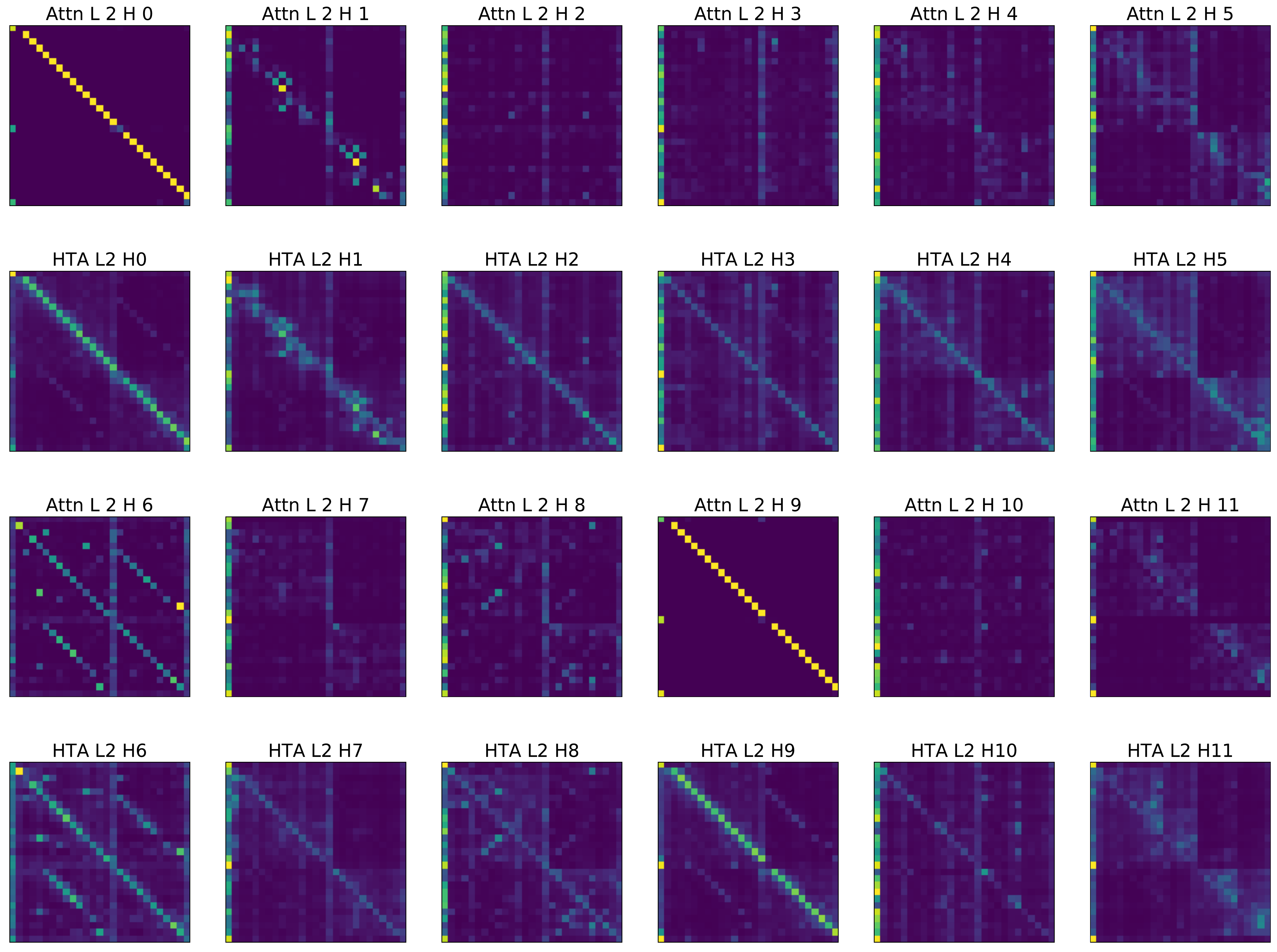}
\caption{Layer 3: rows 1 and 3 represent attention maps, rows 2 and 4 show the corresponding contribution maps.}
\end{figure*}

\begin{figure*}
\centering
\includegraphics[width=0.85\linewidth]{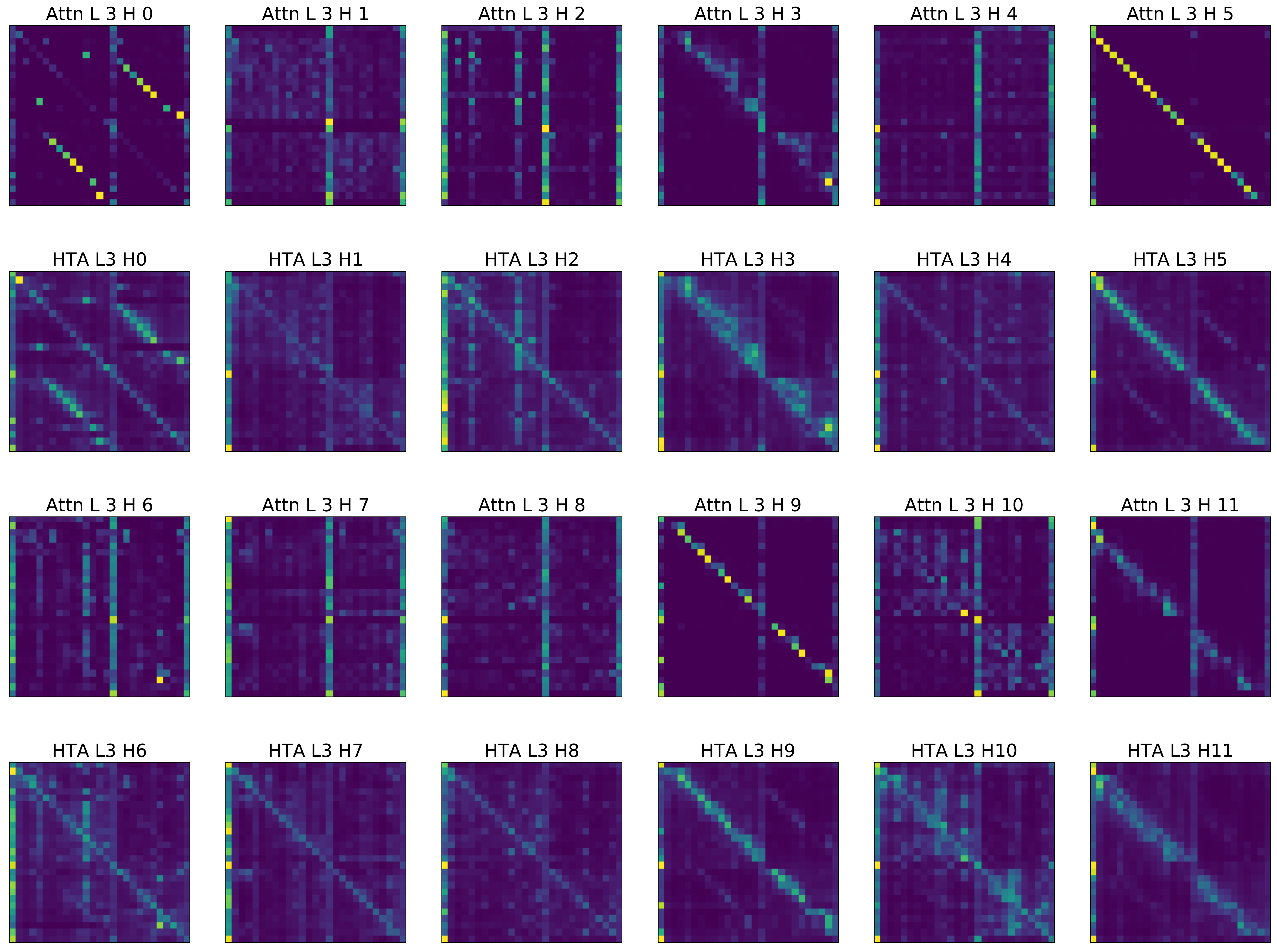}
\caption{Layer 4: rows 1 and 3 represent attention maps, rows 2 and 4 show the corresponding contribution maps.}
\end{figure*}

\begin{figure*}
\centering
\includegraphics[width=0.85\linewidth]{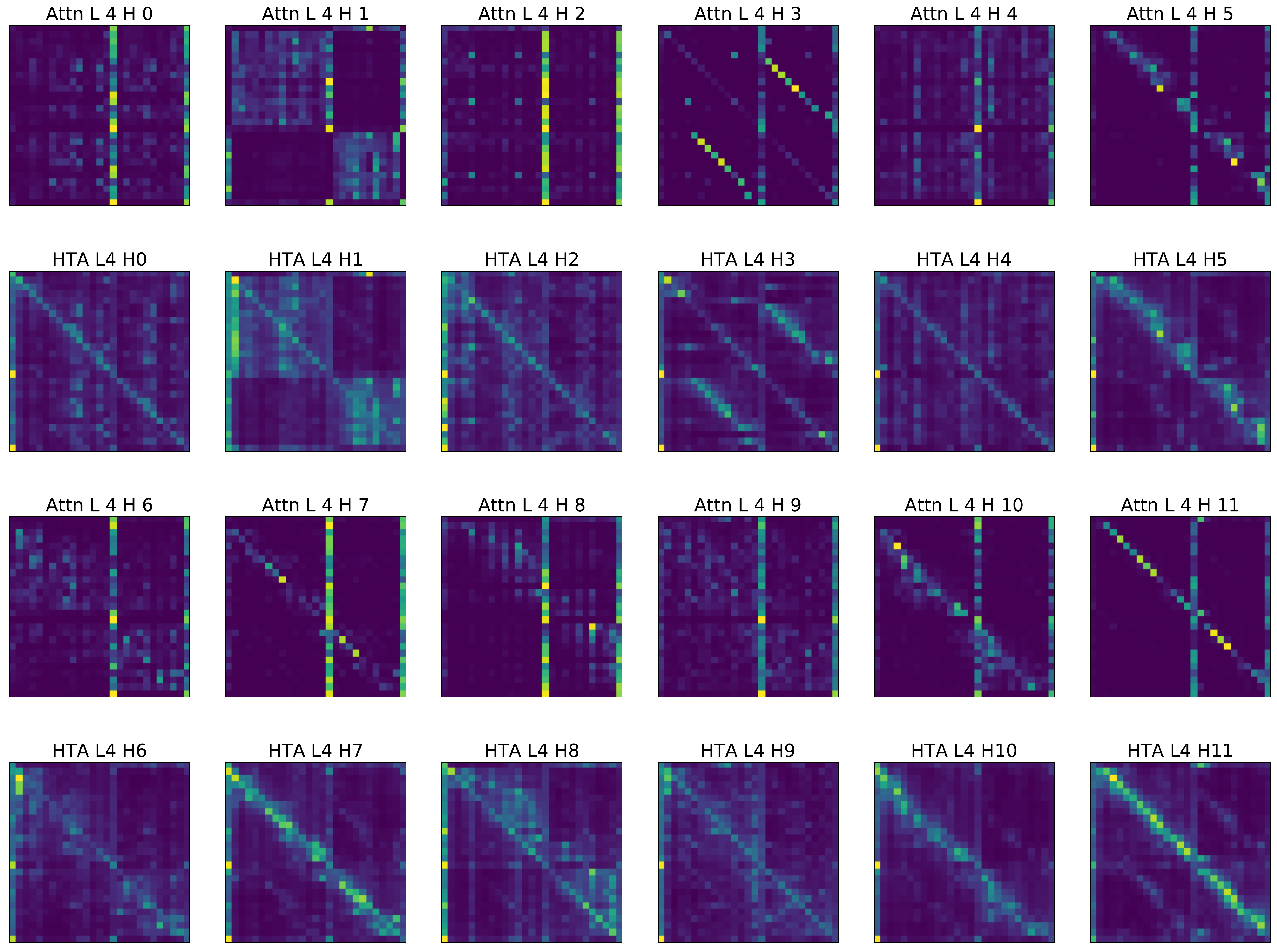}
\caption{Layer 5: rows 1 and 3 represent attention maps, rows 2 and 4 show the corresponding contribution maps.}
\end{figure*}

\begin{figure*}
\centering
\includegraphics[width=0.85\linewidth]{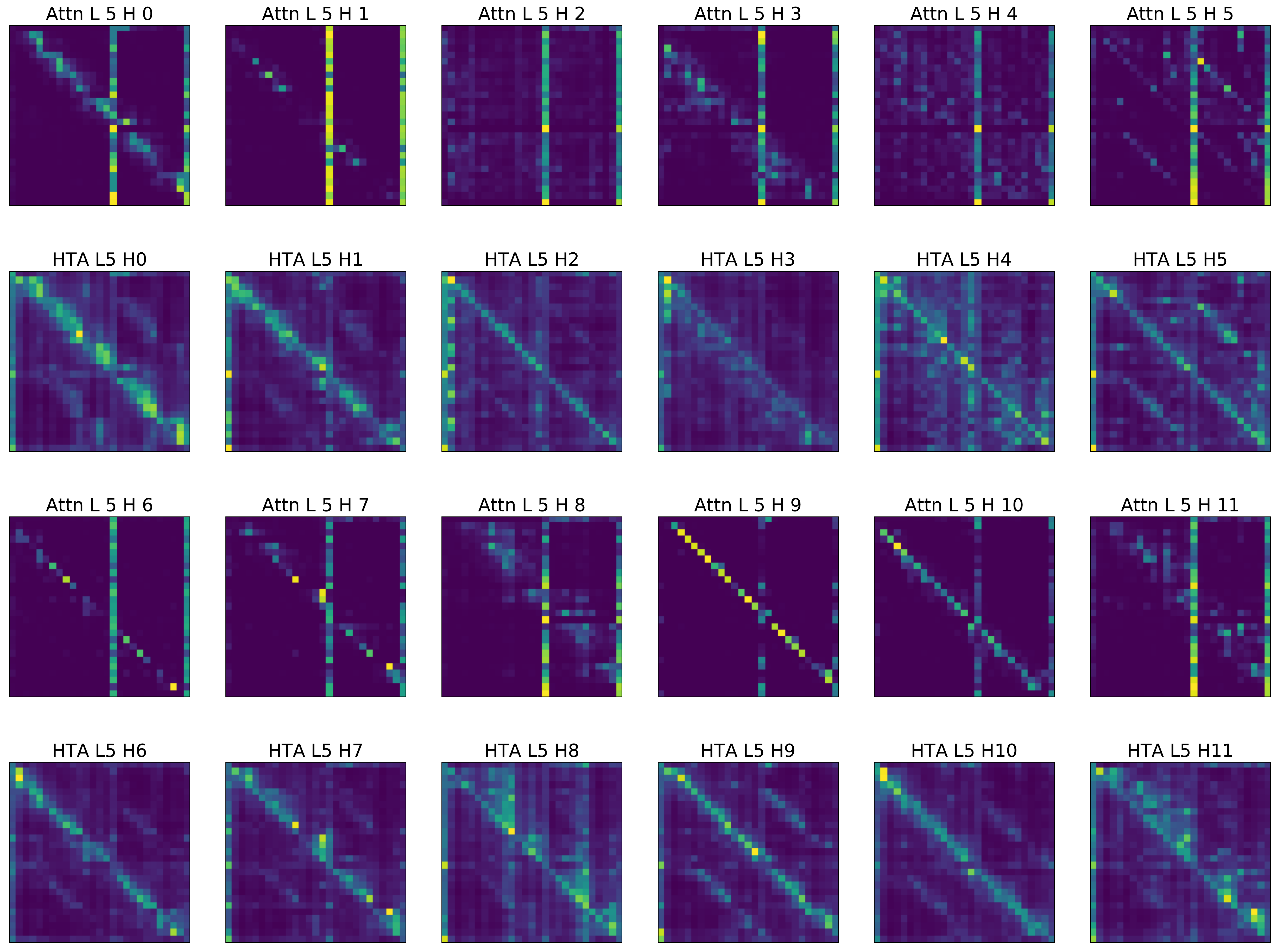}
\caption{Layer 6: rows 1 and 3 represent attention maps, rows 2 and 4 show the corresponding contribution maps.}
\end{figure*}

\begin{figure*}
\centering
\includegraphics[width=0.85\linewidth]{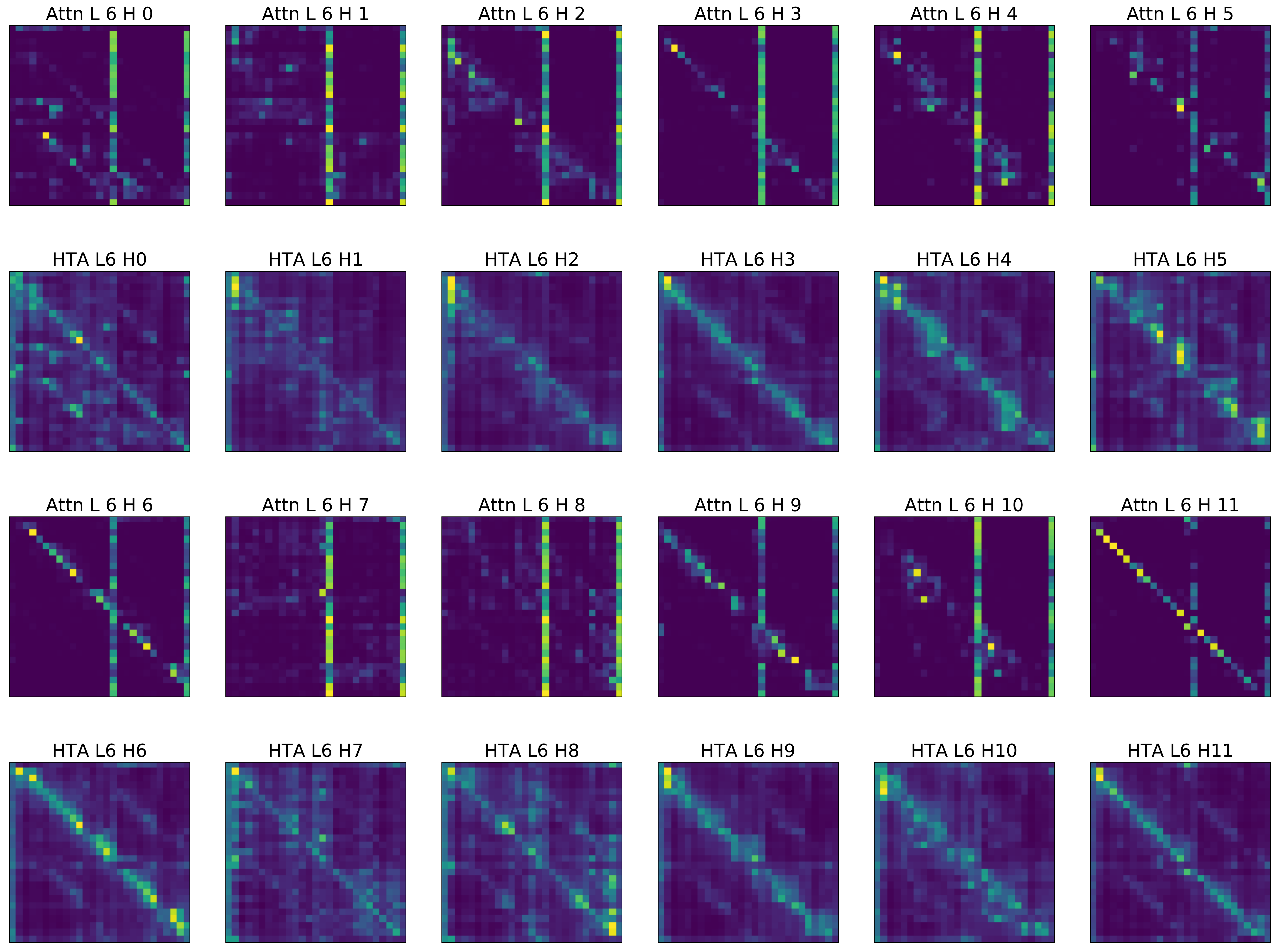}
\caption{Layer 7: rows 1 and 3 represent attention maps, rows 2 and 4 show the corresponding contribution maps.}
\end{figure*}

\begin{figure*}
\centering
\includegraphics[width=0.85\linewidth]{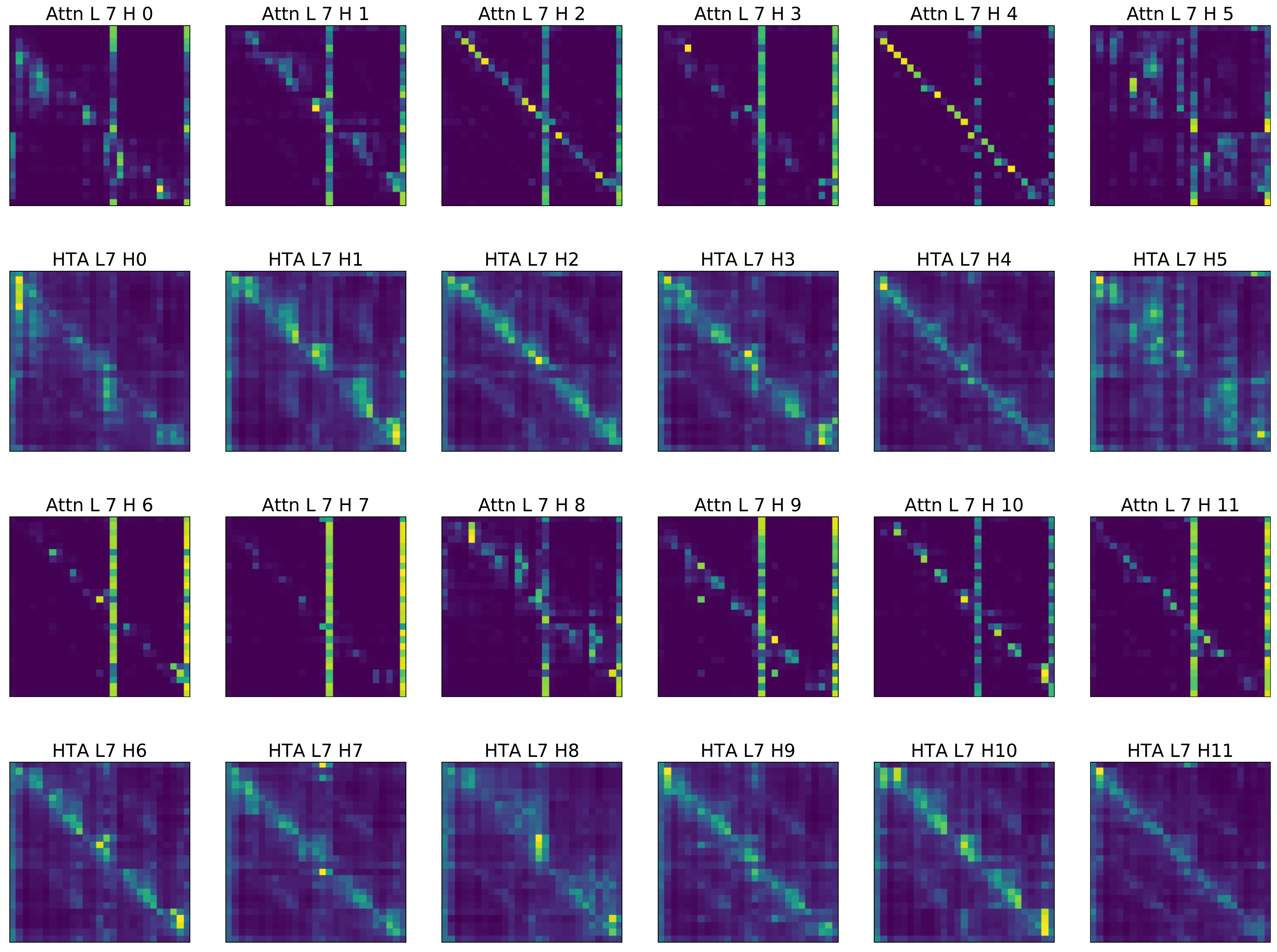}
\caption{Layer 8: rows 1 and 3 represent attention maps, rows 2 and 4 show the corresponding contribution maps.}
\end{figure*}

\begin{figure*}
\centering
\includegraphics[width=0.85\linewidth]{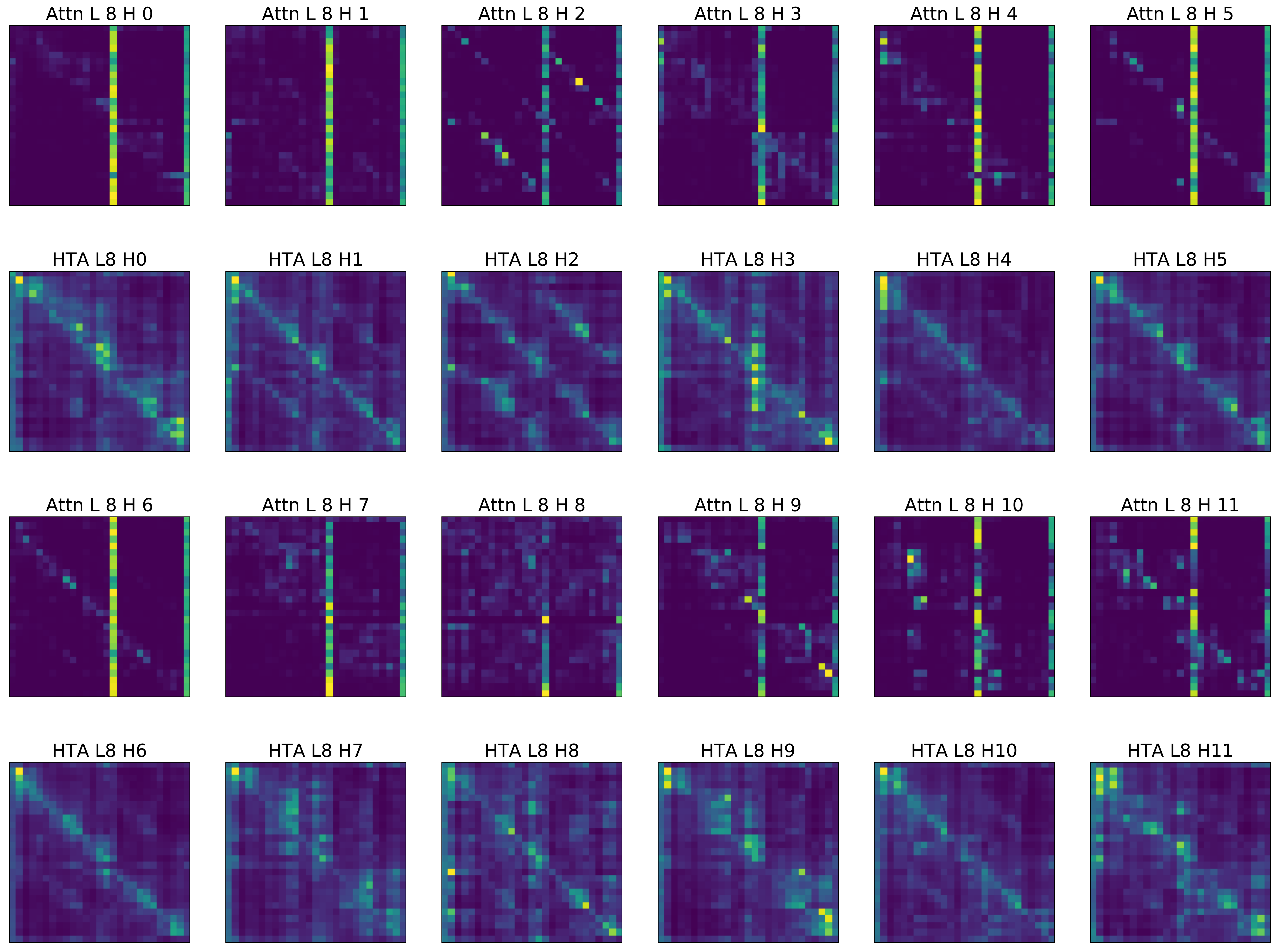}
\caption{Layer 9: rows 1 and 3 represent attention maps, rows 2 and 4 show the corresponding contribution maps.}
\end{figure*}

\begin{figure*}
\centering
\includegraphics[width=0.85\linewidth]{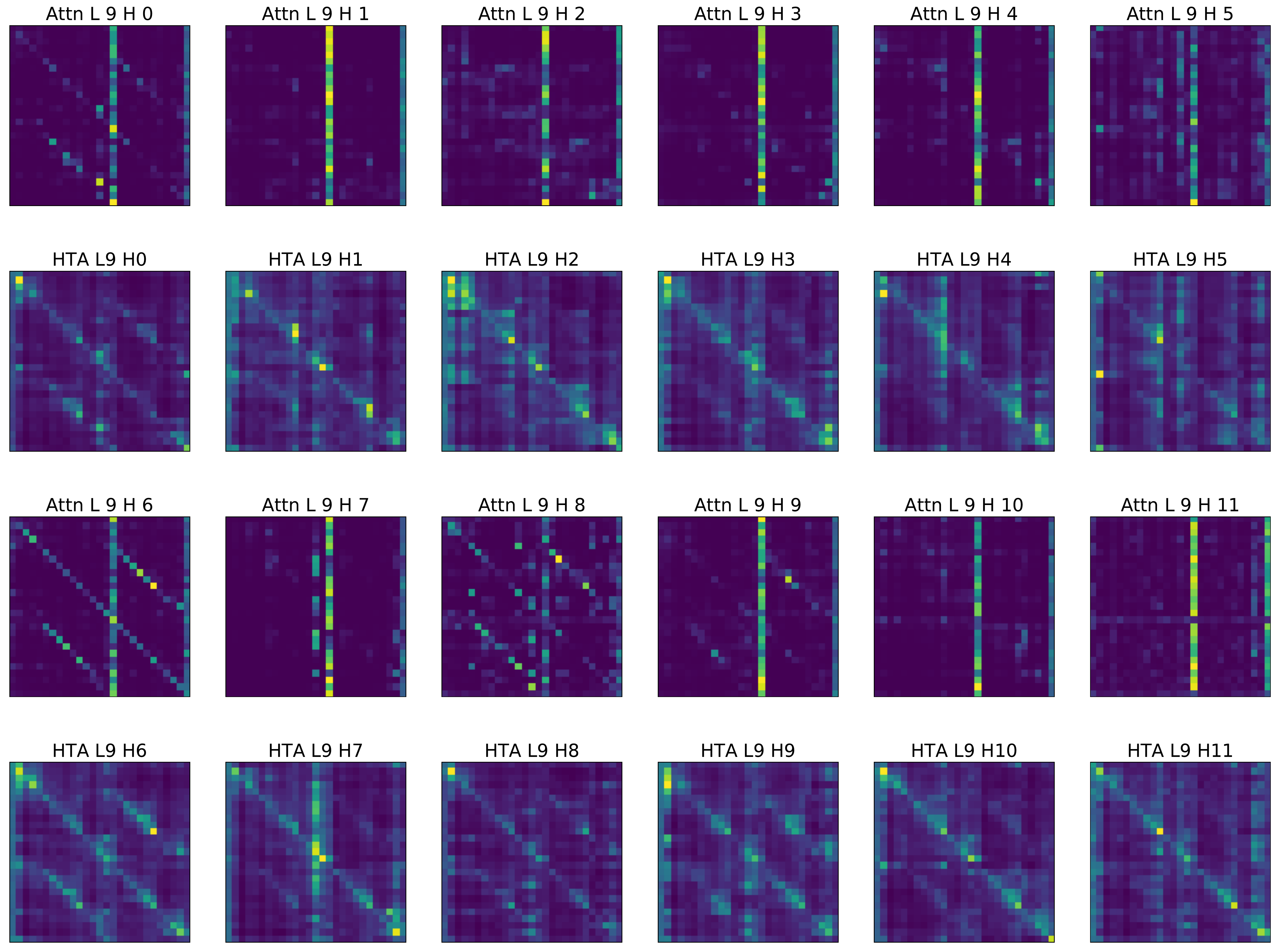}
\caption{Layer 10: rows 1 and 3 represent attention maps, rows 2 and 4 show the corresponding contribution maps.}
\end{figure*}

\begin{figure*}
\centering
\includegraphics[width=0.85\linewidth]{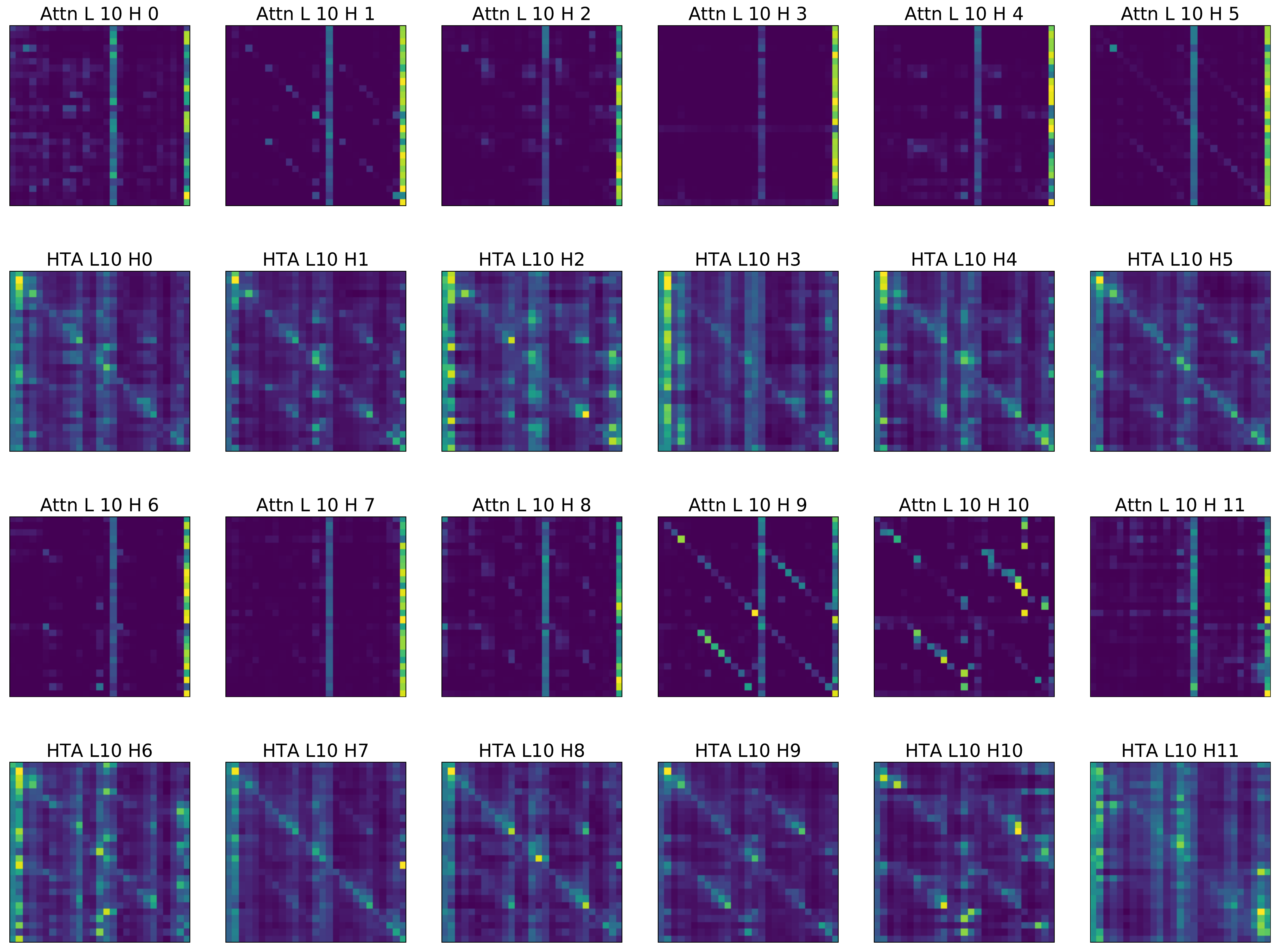}
\caption{Layer 11: rows 1 and 3 represent attention maps, rows 2 and 4 show the corresponding contribution maps.}
\end{figure*}

\begin{figure*}
\centering
\includegraphics[width=0.85\linewidth]{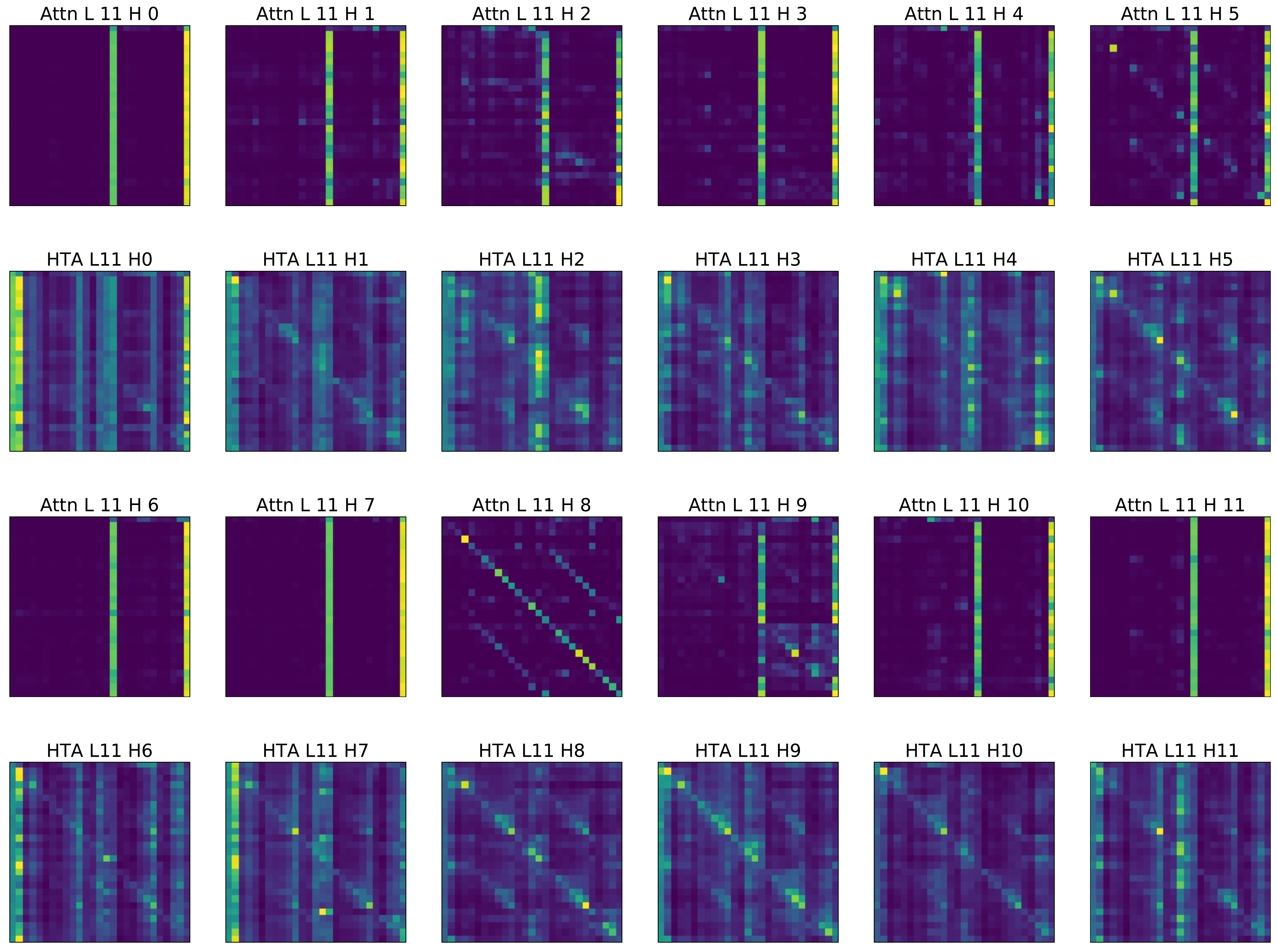}
\caption{Layer 12: rows 1 and 3 represent attention maps, rows 2 and 4 show the corresponding contribution maps.}
\end{figure*}

\end{document}